\documentclass[letterpaper]{article} 
\usepackage{aaai2026}  
\usepackage{times}  
\usepackage{helvet}  
\usepackage{courier}  
\usepackage[hyphens]{url}  
\usepackage{graphicx} 
\usepackage{natbib}  
\usepackage{caption} 
\usepackage{algorithm}
\usepackage{algorithmic}

\usepackage{newfloat}
\usepackage{listings}
\DeclareCaptionStyle{ruled}{labelfont=normalfont,labelsep=colon,strut=off} 
\floatstyle{ruled}
\newfloat{listing}{tb}{lst}{}
\floatname{listing}{Listing}
\usepackage{booktabs}
\usepackage{multirow}
\usepackage{amsmath}
\usepackage{makecell} 
\usepackage{tabularx}
\usepackage{amssymb}
\usepackage{pifont}
\usepackage{array}    
\usepackage{ragged2e}
\usepackage{subcaption}
\usepackage{wasysym}   
\usepackage{enumitem}
\usepackage{cuted}
\nocopyright 

\title{What to Ask Next? Probing the Imaginative Reasoning of LLMs with TurtleSoup Puzzles}

\author {
    Mengtao Zhou\textsuperscript{\rm 1}\thanks{Equal Contribution. Correspondence: bang.liu@umontreal.ca},
    Sifan Wu\textsuperscript{\rm 2}\footnotemark[\value{footnote}],
    Huan Zhang\textsuperscript{\rm 2},
    Qi Sima\textsuperscript{\rm 1},
    Bang Liu\textsuperscript{\rm 2}
}
\affiliations {
    \textsuperscript{\rm 1}Huazhong University of Science and Technology\\
    \textsuperscript{\rm 2}University of Montreal \& Mila
}

\begin{document}

\setlist{nosep}
\setlength{\textfloatsep}{10pt plus 1.0pt minus 2.0pt}
\setlength{\intextsep}{10pt plus 1.0pt minus 2.0pt}
\setlength{\abovecaptionskip}{5pt}
\setlength{\belowcaptionskip}{0pt}

\maketitle
\begin{strip}
    \centering
    \includegraphics[width=1\linewidth]{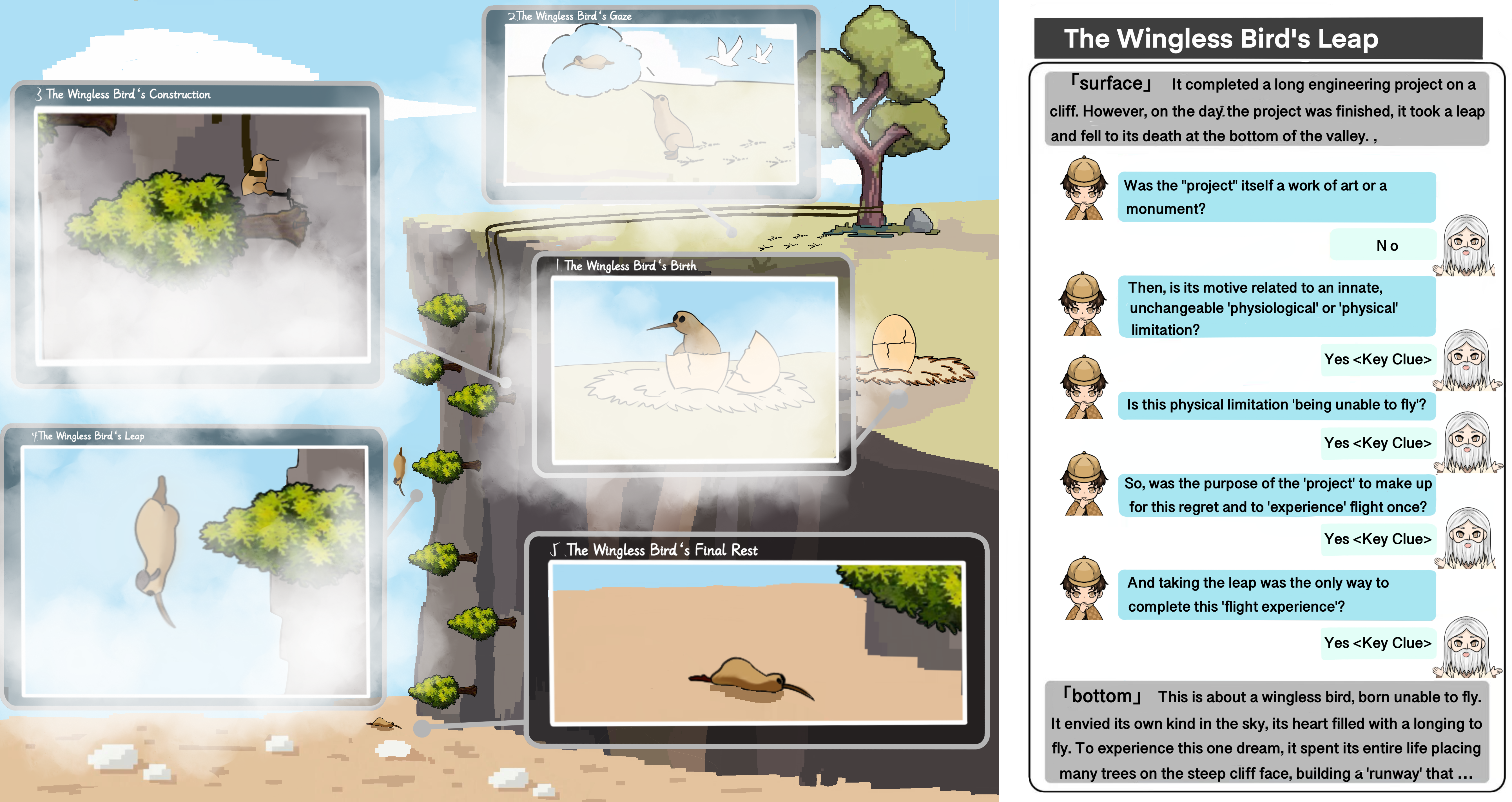}
    \captionof{figure}{Left is the story of ``The Wingless Bird's Leap'' from TurtleSoup-Bench and right is the automatic evaluation through our Mosaic-Agent.}
    \label{fig:enter-label}
\end{strip}

\begin{abstract}
We investigate the capacity of Large Language Models (LLMs) for \textbf{imaginative reasoning}—the proactive construction, testing, and revision of hypotheses in information-sparse environments. Existing benchmarks, often static or focused on social deduction, fail to capture the dynamic, exploratory nature of this reasoning process. To address this gap, we introduce a comprehensive research framework based on the classic ``Turtle Soup'' game, integrating a benchmark, an agent, and an evaluation protocol.
We present \textit{TurtleSoup-Bench}, the first large-scale, bilingual, interactive benchmark for imaginative reasoning, comprising 800 turtle soup puzzles sourced from both the Internet and expert authors. We also propose \textit{Mosaic-Agent}, a novel agent designed to assess LLMs' performance in this setting. To evaluate reasoning quality, we develop a multi-dimensional protocol measuring logical consistency, detail completion, and conclusion alignment.
Experiments with leading LLMs reveal clear capability limits, common failure patterns, and a significant performance gap compared to humans. Our work offers new insights into LLMs' imaginative reasoning and establishes a foundation for future research on exploratory agent behavior.
\end{abstract}


\section{Introduction}
Large Language Models (LLMs)  increasingly serve as the cognitive core of autonomous agents, enabling advanced reasoning, understanding, and decision-making \citep{bubeck2023sparks, xi2023rise,hong2024metagpt,park2023generative, wu2025seeing, Wu2025eccv, wu2023cadllm_workshop}. Yet these gains largely assume \emph{information-complete} settings with fully specified rules, goals, and context. Many real-world tasks are \emph{dynamic} and \emph{information-scarce}—e.g., an archaeologist inferring daily life from a few pottery shards, or a police officer reconstructing a crime from sparse, ambiguous clues.
In such cases, progress depends less on retrieving known facts than on constructing, testing, and revising speculative explanations of the missing pieces. We refer to this advanced reasoning capability as \textbf{imaginative reasoning}.

Prior evaluation of imaginative reasoning in LLMs have relied on multi‑agent social deception games such as Werewolf or Avalon \citep{xu2024,lan-etal-2024}. These works mainly emphasise role inference under hidden information. 
TurtleBench \citep{yu2024turtlebench} employs an evaluation based on static question-answer pairs to assess a model's ability to answer deductive questions but fails to evaluate an agent's core decision-making capability to autonomously and strategically decide what to ask next. While valuable, these benchmarks do not directly measure the iterative process of hypothesis generation, testing, and belief updating that constitutes imaginative reasoning. Their focus remains on social deduction or static knowledge retrieval, not on the agent's ability to creatively explore an unknown problem space.

To bridge this gap and rigorously examine the imaginative reasoning potential of modern LLMs, we leverage the classic narrative‑reasoning game \textbf{Turtle Soup}\footnote{Also known as \emph{Situation Puzzles} or \emph{Yes/No Puzzles}.}. As illustrated in Figure~\ref{fig:enter-label}, each puzzle reveals only a terse, enigmatic scenario as the \emph{soup surface}.  The solver's goal is to recover the complete latent story as the \emph{soup bottom}—by asking a series of yes/no questions. Solving a puzzle, naturally, requires an iterative loop of abductive and deductive logic.Accordingly, we present \textit{TurtleSoup-Bench}, a comprehensive benchmark grounded in turtle soup puzzles, crafted specifically to evaluate the imaginative reasoning ability of LLMs. It encompasses 800 stories drawn from online sources and expert creations. Table~\ref{tab:combined_comparison_and_stats} presents the primary characteristics of our benchmark and contrasts it with prior benchmarks.

\begin{table*}[t]
\small
\centering 

\begin{subtable}[b]{.62\textwidth}
    \centering 
    \renewcommand{\arraystretch}{0.9} 
    \setlength{\tabcolsep}{4pt} 
    
    \begin{tabular}{lccccc}
    \toprule 
    \thead[l]{Benchmark /\\ Framework} & 
    \thead{Inter-\\action} & 
    \thead{Indi-\\vidual} & 
    \thead{Imagin-\\ative} & 
    \thead{Environ-\\ment} & 
    \thead{Evalu-\\ation} \\
    \midrule
    Werewolf      & $\checkmark$ & $\checkmark$ & $\times$ & Adversarial    & Outcome-based \\
    Avalon     & $\checkmark$ & $\checkmark$ & $\times$ & Adversarial    & Outcome-based \\
    Jubensha      & $\checkmark$ & $\checkmark$ & $\times$ & Adversarial    & Outcome-based \\
    MIRAGE       & $\checkmark$ & $\checkmark$ & $\times$ & Cooperative    & Outcome-based \\
    RoleLLM    & $\checkmark$ & $\checkmark$ & $\times$ & Conversational & Output Fidelity \\
    RPGBENCH       & $\checkmark$ & $\checkmark$ & $\times$ & Simulative     & Output Quality \\
    Word Guess  & $\checkmark$ & $\times$     & $\times$ & Adversarial    & Outcome-based \\
    TurtleBench    & $\times$     & $\checkmark$ & $\times$ & Static Puzzle  & Outcome-based \\
    \midrule 
    \textbf{Ours}
                & $\checkmark$ & $\checkmark$ & $\checkmark$ & Non-Advers. & Path \& Fidelity \\
    \bottomrule
    \end{tabular}
    \caption{Comparative Analysis} 
    \label{tab:comparison}
\end{subtable}%
\hfill 
\begin{subtable}[b]{.36\textwidth}
    \centering
    \renewcommand{\arraystretch}{0.9} 
    \setlength{\tabcolsep}{3pt}
    
    \begin{tabular}{l r} 
    \toprule 
    \thead[l]{\textit{TurtleSoup-Bench} Feature} & \thead{Value} \\
    \midrule 
    
    Crime Thriller                & 164 \\
    Mind Game                     & 120 \\
    Supernatural                  & 62  \\
    Constant Change               & 116 \\
    Clever Logic                  & 138 \\
    Original (Expert-Authored)    & 200 \\
    Total                         & 800 \\
    \midrule 
    
    Surface Tokens (per scenario)  & 49.2 \\
    Bottom Tokens (per scenario) & 143.3 \\
    Key Clues (per scenario)       & 5.7 \\
    
    \bottomrule
    \end{tabular}
    \caption{\textit{TurtleSoup-Bench} Statistics}
    \label{tab:statistics}
\end{subtable}
\caption{Comparative analysis of TurtleSoup-Bench against other frameworks and detailed statistics of our benchmark.}
\label{tab:combined_comparison_and_stats}
\end{table*}

Building on TurtleSoup-Bench, we develop the \textbf{Mosaic-Agent} framework to model the iterative process of imaginative reasoning. The framework comprises a Questioner agent, a Responder agent, and a memory module.
Nevertheless, evaluating such a creative, exploratory process presents a significant challenge, as conventional NLP metrics like BLEU or ROUGE collapse multiple dimensions of quality into a single, surface-level similarity score \citep{liu-etal-2023-g}. As established in related fields like creative story generation, a comprehensive assessment requires a multi-faceted approach \citep{yang2024makesgoodstorymeasure}. To address this, our evaluation protocol is designed to disentangle an agent's performance into two distinct perspectives: the quality of the reasoning process and the accuracy of the final result. We evaluate the process using two metrics—Logic Accuracy to assess the coherence of the causal chain and Detail Fidelity to measure factual grounding—while the result is evaluated using Conclusion Match, which holistically compares the agent's final summary against the ground truth. Together, these metrics provide a more granular and meaningful assessment than a monolithic score, offering a finer-grained diagnostic lens for multi-step reasoning systems.

Overall, our main contributions are as follows:
\begin{itemize}
    \item We propose to use TurtleSoup Puzzles as the environments to assess the imaginative reasoning ability of LLMs. To this end, we construct a TurtleSoup-Bench containing 800 scenarios.
    \item We design a multi-agent framework Mosaic-Agent, which aim to solve TurtleSoup puzzle fully automatically by two dynamic interaction agents.
    \item Experiments conducted on TurtleSoup-Bench demonstrate the existing state-of-the-art LLMs struggle in incomplete information scenarios and complex imaginative reasoning tasks.
\end{itemize}


\section{Related Work}

\paragraph{LLM-based Autonomous Agents.}
Building on foundational frameworks like ReAct \citep{yao2023react} and Reflexion \citep{shinn2023reflexion}, recent work has increasingly tested LLM agents in complex multi-agent games. These studies span social deduction games like ``Avalon'' \citep{lan-etal-2024}, ``The Traitors'' \citep{curvo2025}, and ``Werewolf'' \citep{sato-etal-2024, zhang2025dvm, xu2025learning}, to narrative mysteries like ``Murder Mystery Games'' \citep{Qinglin2024, cai2025}, ``SpyGame'' \citep{liang2023, wei2025}, and ``Jubensha'' \citep{wu-etal-2024}. However, these works primarily focus on social strategy and inter-agent confrontation, rather than individual cognition in information-scarce environments. Our work takes a different path by avoiding social dynamics to focus on this fundamental dimension of cognition. We place the agent in a non-adversarial, puzzle-solving context, where the core challenge is to perform imaginative reasoning by constructing a coherent narrative from sparse clues.

\paragraph{Evaluating Reasoning in Interactive Environments.}
The evaluation of LLMs is shifting from static benchmarks like MMLU \citep{hendrycks2021measuring} to dynamic, interactive environments, as static tests cannot assess the procedural capabilities required for active exploration \citep{hsia-etal-2024, eriksson2025}. Consequently, a new generation of interactive benchmarks has emerged, covering areas from open-world exploration \citep{wang2024voyager, yang2025embodiedbench} and multi-turn dialogue games \citep{Ma2024, bai-etal-2024-mt, li2025} to script-based role-playing \citep{wang-etal-2024, yu2025, wang-etal-2025}. The work most adjacent to ours is TurtleBench \citep{yu2024turtlebench}, which pioneered using turtle soup puzzles for reasoning assessment. However, its static question-answering protocol is verification-based, not exploratory, and thus cannot assess the core dynamic process of an agent formulating, testing, and revising beliefs through interaction. Our framework addresses this gap by shifting the evaluation focus from static outcome verification to measuring the entire dynamic process of imaginative reasoning.

\section{\textit{TurtleSoup-Bench}: A Benchmark for Imaginative Reasoning}

\paragraph{Data Collection and Authoring.}
To balance the benchmark's breadth and novelty, we adopted a dual-source data collection strategy. First, we collected a large number of Chinese turtle soup puzzles from well-known online puzzle communities (e.g., \citep{TangWebsite}). To select high-quality samples from this extensive pool, we implemented a two-stage filtering process. The first stage was a community-based pre-screening, where we retained only those stories with high upvote counts, top ratings, or a large number of positive comments, ensuring their popularity and recognized quality among players. This step narrowed down the candidates. The second stage was an expert-led final selection, where our expert team manually reviewed the pre-screened stories. Based on criteria such as logical soundness, narrative cleverness, and suitability for LLM evaluation, they selected the final 300 stories. To address the critical issue of data contamination and introduce more challenging scenarios, we then recruited a team of five experienced puzzle design experts to author 100 entirely new stories. The experts followed strict design principles during creation, including logical consistency (the bottom fully explains the premise), non-obviousness (the bottom is clever and non-intuitive), and self-containment (solvable only via ``Yes/No/Unknown'' questions). The 40 to 60 minutes required to craft a single story ensures the originality and high quality of this dataset.

\paragraph{Data Curation and Annotation.}
After compiling the 400 stories, we conducted a rigorous curation and annotation process for all data. All flawed or ambiguous samples identified during the selection process were corrected and re-edited by experts. Subsequently, to support a more nuanced analysis of agent capabilities, our expert team classified the collected stories into five core narrative genres, with our original stories treated as a distinct sixth category. Furthermore, a core contribution of our benchmark is the manual annotation of a \emph{Key Clue Library} $K_{\text{lib}}$ for every story. These expert-defined clues represent pivotal turning points in the reasoning process and provide effective guidance signals for the agent. To extend the benchmark's utility to the global research community, the complete corpus of 400 Chinese stories was professionally translated and culturally adapted into an equivalent set of 400 English stories.Ultimately, our \textit{TurtleSoup-Bench} comprises 800 scenarios, with the detailed distribution shown in Table~\ref{tab:statistics}. Some of the scenarios are shown in the Appendix.

\section{Mosaic-Agent Framework: Interactive Environment for TurtleSoup-Bench}
\begin{figure}[t]
    \centering
    \includegraphics[width=0.47\textwidth]{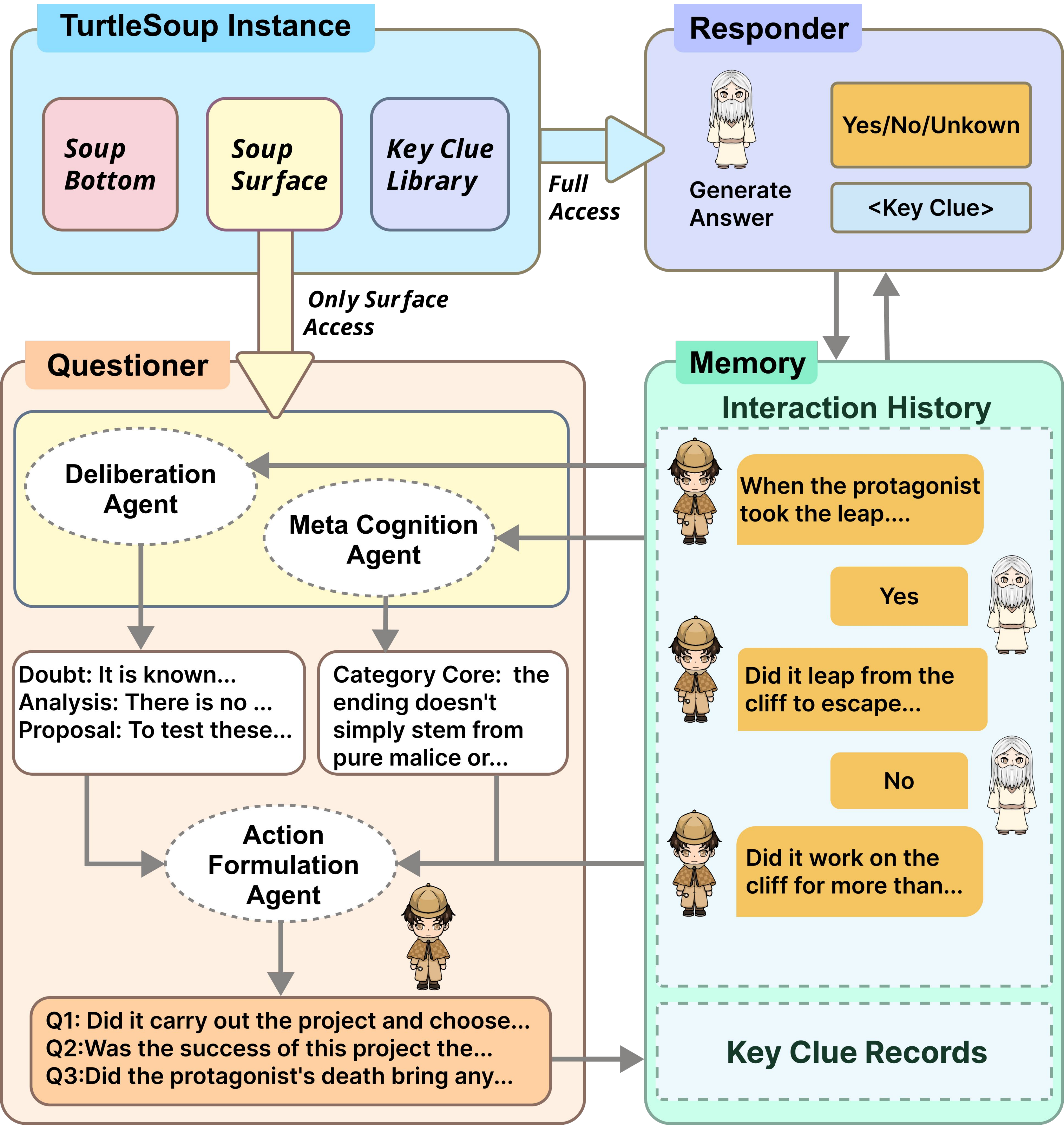}
    \caption{The Mosaic-Agent Framework}
    \label{fig:main_frame}
\end{figure}

This section outlines our framework, Mosaic-Agent, designed to simulate a TurtleSoup puzzle solution. The goal is to find the \emph{soup bottom} behind the \emph{soup surface} by modeling multi-turn interaction between a questioner and a responder agent grounded in the real TurtleSoup situation~\cite{wiki:SituationPuzzle}.
As illustrated in Figure \ref{fig:main_frame}, the framework consists of: the questioner agent, the responder agent, and memory module. The questioner agent aims to act like the player to propose imaginative questions. The responder agent is like god to response the question and hint about key clue. And the memory module acts like a detective's notebook, recording the full conversation and pivotal clues.



\subsection{Questioner: A Deliberative Cognitive Architecture}

Efficient human problem-solving follows a logic progressing from analysis to decision-making. The process begins by analyzing information to form hypotheses \citep{MarcelPANS}, then uses strategic foresight to simulate potential outcomes \citep{Schacter2007}. Ultimately, this analysis leads to a key decision—such as asking a highly informative question—aimed at most effectively reducing uncertainty \citep{Kidd2015-ze}. We leverage this cognitive model by decoupling our questioner agent's question-generation mechanism into three corresponding processes: deliberation, meta-cognition, and action generation, to replicate the efficiency of human thought.

\textbf{Deliberation Agent.}
This agent serves as the core analytical engine. At the beginning of each decision cycle, it processes existing information to form a comprehensive understanding of the situation and identify key directions for the next exploration.
The workflow is designed as a hierarchical cognitive process that integrates two distinct but complementary levels of cognition: one for local analysis for immediate reaction, and another for global synthesis to build a macroscopic understanding. This progressive, local-to-global design is manifested in:

First, the agent performs a quick analysis of the most recent question-answer pair, $(q_{t-1}, a_{t-1})$, where $q$ denotes the question, $a$ the answer, and $t$ the current turn. The purpose of this step is to quickly parse the newly acquired local information from the last interaction, establishing a clear starting point for the next line of reasoning.

To form a more macroscopic and in-depth understanding, the agent periodically (at a fixed interval of $k$ turns) conducts a global deliberation to examine the global picture and prevent cognitive myopia. The core of this process is updating its internal Belief State, $B_t$, a structured json object of the agent's understanding of the story's core logic $L_t$, key details $D_t$, and overall conclusion $C_t$.This update is handled by the function $f_{\text{sum}}$, which directs the agent to retrieve the Interaction History $H_{t-1}$, Key Clue Records $K_{\text{rec}}$ from the shared Memory module, and the belief state from the previous cycle $B_{t-k}$ into the new belief state.
\begin{equation}
B_t = (L_t, D_t, C_t) = f_{\text{sum}}(H_{t-1}, K_{\text{rec}}, B_{t-k})
\end{equation}
Then the agent performs a self-diagnosis on the updated belief state $B_t$ to identify logical gaps or missing information. The function $f_{\text{advice}}$ performs this diagnosis, instructing the agent to generate a structured ``Analysis and Proposal Set'' (APS) based on the current belief state $B_t$ and the \emph{soup surface} $S_{\text{surf}}$:
\begin{equation}
    APS_t = f_{\text{advice}}(B_t, S_{\text{surf}}) = \{ (d_j, an_j, p_j) \}_{j=1}^m
\end{equation}
This set contains $m$ tuples, where each tuple represents a specific doubt $d_j$, its corresponding analysis $an_j$, and a question proposal $p_j$ designed to resolve the doubt. Detailed illustration for are in Appendix. 

\textbf{Meta-cognition Agent.}
This agent dynamically adjusts the agent's macro-level strategy by periodically classifying the puzzle's narrative genre.

A static strategy is brittle in complex exploration. The agent initiates a judgment whenever it acquires several new clues or after several turns with no progress. This judgment involves a three-vote majority classification based on the \emph{soup surface} $S_{\text{surf}}$, Interaction History $H_{t-1}$, and the Key Clue Records $K_{\text{rec}}$, to assign the puzzle to a narrative genre. An occasional misclassification can cause severe strategy oscillation. Imagine the agent misclassifies a crime story as supernatural; it would waste turns asking about ghosts.

To prevent this, we use a Smoothed Confidence mechanism. It maintains a policy confidence, $c$, representing the agent's confidence in its current assessment, initialized to a neutral 0.5 at the start of the game. When a new vote yields a confidence score $v_c$, we update it via the Exponential Moving Average formula:
\begin{equation}
    c_t^{\text{smooth}} = \alpha \cdot c_{t-1} + (1-\alpha) \cdot v_c
\end{equation}
where $c_{t-1}$ is the confidence from the previous cycle. A key stability feature is that the agent switches its strategy only if the new smoothed confidence, $c_t^{\text{smooth}}$, exceeds the sum of the old confidence and a predefined threshold, $\tau_{\text{switch}}$. In our study, $\alpha$ is set to 0.7, and $\tau_{\text{switch}}$ is set to 0.1.

A successful genre switch makes the agent adopt a new questioning strategy, We designed unique questioning strategies for each narrative genre in collaboration with human experts, following the principles of compositional reasoning \citep{press-etal-2023-measuring}. These strategies target the typical logical structures of each genre to make subsequent questioning more focused.

\textbf{Action Formulation agent.}
In this agent, we aim to integrate the various cognitive outputs to formulate the final single question action $q_t$.

First, in the candidate generation stage, the agent combines the Analysis and Proposal Set $APS_t$ from the Deliberation agent with the chosen questioning strategy from the Meta-cognition agent to generate three candidate questions, $Q_{\text{cand}}$.
Next, in the optimal screening stage, we let the agent act as its own decision-making critic. The agent considers the full Interaction History $H_{t-1}$, its own analysis $APS_t$, and a blacklist, $\mathbb{B}$. This blacklist is created at the start of the game and is dynamically updated with questions that are identified as ineffective or redundant during gameplay. Based on this complete context, the agent selects from the candidates the single best question, $q_t$, that is most likely to yield new information while avoiding redundancy:
\begin{equation}
q_t = \underset{q' \in Q_{\text{cand}}}{\operatorname{argmax}} \text{ Score}(q' | H_{t-1}, APS_t, \mathbb{B})
\end{equation}
This complete Deliberation-MetaCognition-Action process ensures that every question posed by the agent is well-considered and most likely to lead to the truth. Finally, $q_t$ is sent to the Memory module and passed to the Responder.


\subsection{Responder: The God as an Interactive Environment}

In our framework, the Responder is not a game-playing adversary to be evaluated, but is designed as a deterministic and truthful interactive environment, acting as the ``God''. Its core function, $f_{\text{respond}}$, maps the Questioner's question at turn $t$, $q_t$, to a feedback tuple $(a_t, f_t)$, based on the \emph{soup bottom} $S_{\text{bot}}$ and the \emph{Key Clue Library} $K_{\text{lib}}$:
\begin{equation}
    (a_t, f_t) = f_{\text{respond}}(q_t, S_{\text{sol}}, K_{\text{tips}})
\end{equation}
This tuple consists of two components, the answer $a_t$ and the key clue flag $f_t$, which are detailed as follows:

\textbf{Answer Generation.}
Given the Questioner's query $q_t$ and the \emph{soup bottom} $S_{\text{bot}}$, the agent first makes a logical judgment as to whether the content of $q_t$ is consistent with $S_{\text{bot}}$. Based on this judgment, the agent provides a standardized answer, $a_t$, strictly confined to one of three categories:
\begin{itemize}
    \item \textbf{Yes}: Indicates the statement in the question is true according to the solution. For instance, if the solution is ``The killer wore a red coat'', the answer to ``Did the killer wear a red coat?'' is ``Yes''.
    \item \textbf{No}: Indicates false statement. For the same example, the answer to ``Did the killer wear a blue coat?'' is ``No''.
    \item \textbf{Unknown}: This response serves a dual purpose, encompassing cases where the information is either genuinely absent from the solution or simply irrelevant to the core mystery. For simplicity, we group these two cases into a single ``Unknown'' label, as both represent information that does not advance the puzzle's solution.
\end{itemize}

\textbf{Key Clue Identification.}
To better guide the Questioner's reasoning and help it recognize breakthroughs, we introduce the boolean flag $f_t$. That is, not all ``Yes/No'' answers are equally important, and the questions that hit upon the core of the puzzle will be highlighted by this flag.

The agent determines if the semantics of the question $q_t$ are directly relevant to any of the predefined key clues in $K_{\text{lib}}$. If they are, $f_t$ is set to true, and a \texttt{<Key Clue>} marker is appended to the answer string $a_t$. And they are passed to the Questioner via the memory module. For example, if the core of a solution is ``The man was a dwarf and couldn't reach the button,'' a key clue in $K_{\text{lib}}$ might be ``The man's height is a critical factor.'' When the Questioner asks, ``Was the man short?'', the answer would be ``Yes'' and $f_t$ would be true, so the final complete answer returned to the questioner would be the string ``Yes\texttt{<Key Clue>}''.


\begin{table*}[t]
\centering
\renewcommand{\arraystretch}{0.9} 

\setlength{\tabcolsep}{3pt} 

\begin{tabular}{@{}l c *{6}{c@{ }c}@{}}
\toprule
\textbf{Model} & \textbf{Lang} & \multicolumn{2}{c}{\textbf{Crime Thrill.}} & \multicolumn{2}{c}{\textbf{Mind Game}} & \multicolumn{2}{c}{\textbf{Supernat.}} & \multicolumn{2}{c}{\textbf{Const. Change}} & \multicolumn{2}{c}{\textbf{Clev. Logic}} & \multicolumn{2}{c}{\textbf{Orig. Data.}} \\
\cmidrule(lr){3-4} \cmidrule(lr){5-6} \cmidrule(lr){7-8} \cmidrule(lr){9-10} \cmidrule(lr){11-12} \cmidrule(lr){13-14}
& & Score & Reduce & Score & Reduce & Score & Reduce & Score & Reduce & Score & Reduce & Score & Reduce \\
\midrule
\multirow{2}{*}{claude-3.7-sonnet} 
& Ch & \textbf{54.54} & (-15.34) & \textbf{54.66} & (-18.39) & \textbf{58.67} & (-16.39) & \textbf{63.19} & (-8.93)  & \textbf{61.51} & (-11.82) & 56.82 & (-11.05) \\
& En & 28.18 & (-41.70) & 30.63 & (-42.42) & \textbf{37.00} & (-38.06) & 39.93 & (-32.19) & 39.97 & (-33.36) & 31.56 & (-36.31) \\
\cmidrule{1-14}
\multirow{2}{*}{gemini-2.5-flash}  
& Ch & 52.39 & (-17.49) & 47.93 & (-25.12) & 57.42 & (-17.64) & 62.90 & (-9.22)  & 45.45 & (-27.88) & 51.74 & (-16.13) \\
& En & \textbf{33.59} & (-36.29) & \textbf{30.73} & (-42.32) & 35.00 & (-40.06) & 40.95 & (-31.17) & \textbf{42.17} & (-31.16) & \textbf{34.15} & (-33.72) \\
\cmidrule{1-14}
\multirow{2}{*}{deepseek-r1}       
& Ch & 46.94 & (-22.94) & 46.88 & (-26.17) & 53.61 & (-21.45) & 59.36 & (-12.76) & 57.51 & (-15.82) & \textbf{57.14} & (-10.73) \\
& En & 29.82 & (-40.06) & 24.35 & (-48.70) & 36.77 & (-38.29) & 41.40 & (-30.72) & 36.86 & (-36.47) & 29.76 & (-38.11) \\
\cmidrule{1-14}
\multirow{2}{*}{gpt-4o}            
& Ch & 28.79 & (-41.09) & 28.57 & (-44.48) & 32.26 & (-42.80) & 38.07 & (-34.05) & 34.38 & (-38.95) & 33.85 & (-34.02) \\
& En & 31.46 & (-38.42) & 26.78 & (-46.27) & 32.19 & (-42.87) & \textbf{41.76} & (-30.36) & 39.23 & (-34.10) & 30.68 & (-37.19) \\
\cmidrule{1-14}
\multirow{2}{*}{qwen3-32b}         
& Ch & 42.07 & (-27.81) & 39.62 & (-33.43) & 52.19 & (-22.87) & 54.84 & (-17.28) & 46.58 & (-26.75) & 48.55 & (-19.32) \\
& En & 15.66 & (-54.22) & 14.67 & (-58.38) & 17.45 & (-57.61) & 22.95 & (-49.17) & 21.38 & (-51.95) & 20.25 & (-47.62) \\
\cmidrule{1-14}
\multirow{2}{*}{llama3-8b-instruct}
& Ch & 10.51 & (-59.37) & 8.95  & (-64.10) & 6.74  & (-68.32) & 13.22 & (-58.90) & 12.57 & (-60.76) & 7.70  & (-60.17) \\
& En & 6.57  & (-63.31) & 5.12  & (-67.93) & 7.84  & (-67.22) & 8.67  & (-63.45) & 11.52 & (-61.81) & 6.88  & (-60.99) \\
\cmidrule{1-14}
\multicolumn{2}{@{}l}{\textbf{Human Baseline}} 
     & \textbf{69.88} & & \textbf{73.05} & & \textbf{75.06} & & \textbf{72.12} & & \textbf{73.33} & & \textbf{67.87} & \\
\bottomrule
\end{tabular}
\caption{Overall scores $S_{\text{overall}}$ of different models and the human baseline, grouped by language, across all six genres from \textit{TurtleSoup-Bench}. Scores are presented as percentages, with the reduction from the human baseline noted. The best-performing model in each column for each language group is highlighted in bold.}
\label{tab:main_scores}
\end{table*}

\subsection{Memory Module}
The memory module acts as a central hub in the agent's cognitive loop, recording information that facilitates the questioner's reasoning process. The module is partitioned into two key components: the complete interaction history and a curated record of key clues.

\textbf{Interaction History.}
The Interaction History $H_t$ is a complete, chronological log of every question posed by the Questioner and every answer given by the Responder. Its primary function is to provide complete context to the questioner, allowing agents to reflect on the entire conversational flow and prevent reasoning gaps due to forgotten information.

\textbf{Key Clue Records.}
The Key Clue Records $K_{\text{rec}}$ is a filtered, high-value memory pool. When a response from the Responder is flagged with \texttt{<Key Clue>}, the corresponding question-answer pair ($q_t, a_t$) is stored here. The purpose of this design is to allow the Questioner to quickly locate and access the information that is pivotal to solving the puzzle. 

\subsection{Automated Evaluation Protocol}
\label{subsec:Evaluation}
To objectively evaluate Mosaic-Agent's open-ended summary, our protocol first decomposes the \emph{soup bottom} $S_{\text{bot}}$ into two sets of structured evaluation points using an LLM: Core Logic Points $L_{\text{true}}$ and Key Details $D_{\text{true}}$. The number of points elicited is determined adaptively by a set of heuristic rules based on the solution's length and complexity---for instance, the number of logic points ranges from 2 to 5 depending on text length---to ensure comprehensive coverage.

The Questioner's summary, $B_{\text{final}}=(L_{\text{pred}}, D_{\text{pred}}, C_{\text{pred}})$, is then assessed across three dimensions. The motivation for these dimensions is to disentangle the agent's capabilities from the dual perspectives of the reasoning process and the final result. To evaluate the process, Logic Accuracy $S_{\text{logic}}$ measures the coherence of the causal chain by matching $L_{\text{pred}}$ against $L_{\text{true}}$, while Detail Fidelity $S_{\text{details}}$ measures the factual grounding by matching $D_{\text{pred}}$ against $D_{\text{true}}$. To evaluate the result, Conclusion Match $S_{\text{conclusion}}$ provides a holistic assessment by comparing the final summary text $C_{\text{pred}}$ with the \emph{soup bottom} $S_{\text{bot}}$.

In implementation, we use the latest LLM to perform semantic matching. To align with human judgment, the logic and detail scores undergo a Two-Threshold Calibration: a Validity Threshold of 0.5 ensures rigor by filtering out weakly-related matches, while a High-Confidence Threshold of 0.8 normalizes any strong match to a full score of 1.0, making the evaluation robust to paraphrasing. The final Overall Score is a weighted sum:
\begin{equation}
    S_{\text{overall}} = w_l \cdot S_{\text{logic}} + w_d \cdot S_{\text{details}} + w_c \cdot S_{\text{conclusion}}
\end{equation}
where the weights $w_l, w_d, \text{and } w_c$ are set to 0.3, 0.3, and 0.4 in our study, respectively.

\subsection{Human Performance Baseline}
\label{subsec:human}
To illustrate the difficulty of TurtleBench, we recruited 4 human players with extensive experience in the Turtle Soup game. These experts played all 400 Chinese scenarios in \textit{TurtleSoup-Bench}. We meticulously recorded the gameplay of each human player, including the complete sequence of questions, the number of turns taken to solve each puzzle, and their final summary. Critically, we do not subjectively score the human players. Instead, their generated final summaries are fed into the same Automated Evaluation Protocol described in the Section~\ref{subsec:Evaluation}. This ensures that the performances of both the agent and the human players are compared under an identical, objective standard, guaranteeing a fair comparison. 

\section{Experiments}
\subsection{Experimental Setup}

\textbf{Environment and Models.}
All experiments are conducted on our \textit{TurtleSoup-Bench}. We select a representative suite of state-of-the-art LLMs for evaluation, including claude-3.7-sonnet\cite{anthropic2025}, gemini-2.5-flash\cite{comanici2025gemini}, deepseek-r1\cite{guo2025deepseek}, gpt-4o\cite{hurst2024gpt}, qwen3-32b\cite{yang2025qwen3}, and llama3-8b-instruct\cite{dubey2024llama}. In our symmetric design, the Questioner and Responder employ the same model to simultaneously assess its reasoning and comprehension fidelity. To ensure fairness, we use deepseek-r1 for evaluation.

\textbf{Evaluation Protocol.}
We employ the automated protocol from Section ~\ref{subsec:Evaluation}, with primary metrics being $S_{\text{logic}}$, $S_{\text{details}}$, $S_{\text{conclusion}}$, and $S_{\text{overall}}$. Furthermore, we introduce the Human Baseline established in Section ~\ref{subsec:human} as a reference to measure the performance gap between LLM agents and human experts.

\textbf{Implementation Details.}
In all experiments, the periodic deliberation interval $k$ is set to 5, and the maximum number of question turns $N_{\text{max}}$ is 30. Due to cost, all scenarios in this study are run only once.


\subsection{Quantitative Analysis}
Table~\ref{tab:main_scores} presents the results of our quantitative evaluation on \textit{TurtleSoup-Bench}, providing concrete evidence for analyzing the imaginative reasoning capabilities of LLMs.

The results reveal a clear performance stratification among models that 
\textbf{top-tier proprietary models form a leading group, while open-source models, even the larger-parameter qwen3-32b (48.1\%), exhibit a significant performance gap compared to the former}, such as claude-3.7-sonnet (58.8\%). While the human performance achieved higher score as 67.87\% for least. We believe that the performance bottleneck arises not only from the Questioner's deficiency in generating effective exploratory hypotheses but also from the Responder's difficulty in accurately understanding questions and faithfully providing correct answers. A flawed Responder introduces environmental noise that systematically derails the reasoning process.

Besides, we found that \textbf{model performance correlates strongly with the narrative paradigm}, revealing significant capability biases among different models. A compelling example is gemini-2.5-flash, which scores as high as 62.9\% on ``Constant Change'', yet its performance drops sharply to 45.5\% on ``Clever Logic'', which demands a non-intuitive logical leap.  And when a task requires modeling complex human intent (as in ``Crime Thriller'') or non-linear logical reasoning, the performance of most models is severely challenged. This reveals that current LLM imagination is not a general but a collection of specialized skills optimized for specific tasks, with limited generalization.

Furthermore, \textbf{a systematic performance decline is observed across almost all models on the English dataset}. For example, the average score of deepseek-r1 drops by nearly 40\%. Although we have done some cultural adaptations, the subtleties of many puzzles are rooted in their cultural and linguistic origins. The introduction of ambiguity and semantic loss during cross-language conversion also increase the difficulty of reasoning.

Finally, \textbf{a significant chasm persists between the best-performing agents and the human baseline.} The top model, claude-3.7-sonnet, still lags behind human experts by approximately 13 percentage points. The highly effective intuition, creative hypothesis generation, and the ability to efficiently eliminate irrelevant options from a vast space of possibilities that human players exhibit are core capabilities that current models, reliant on probabilistic pattern-matching, have yet to replicate.

\subsection{Qualitative and Error Analysis}

\begin{figure}[h]
    \centering
    \includegraphics[width=1\linewidth]{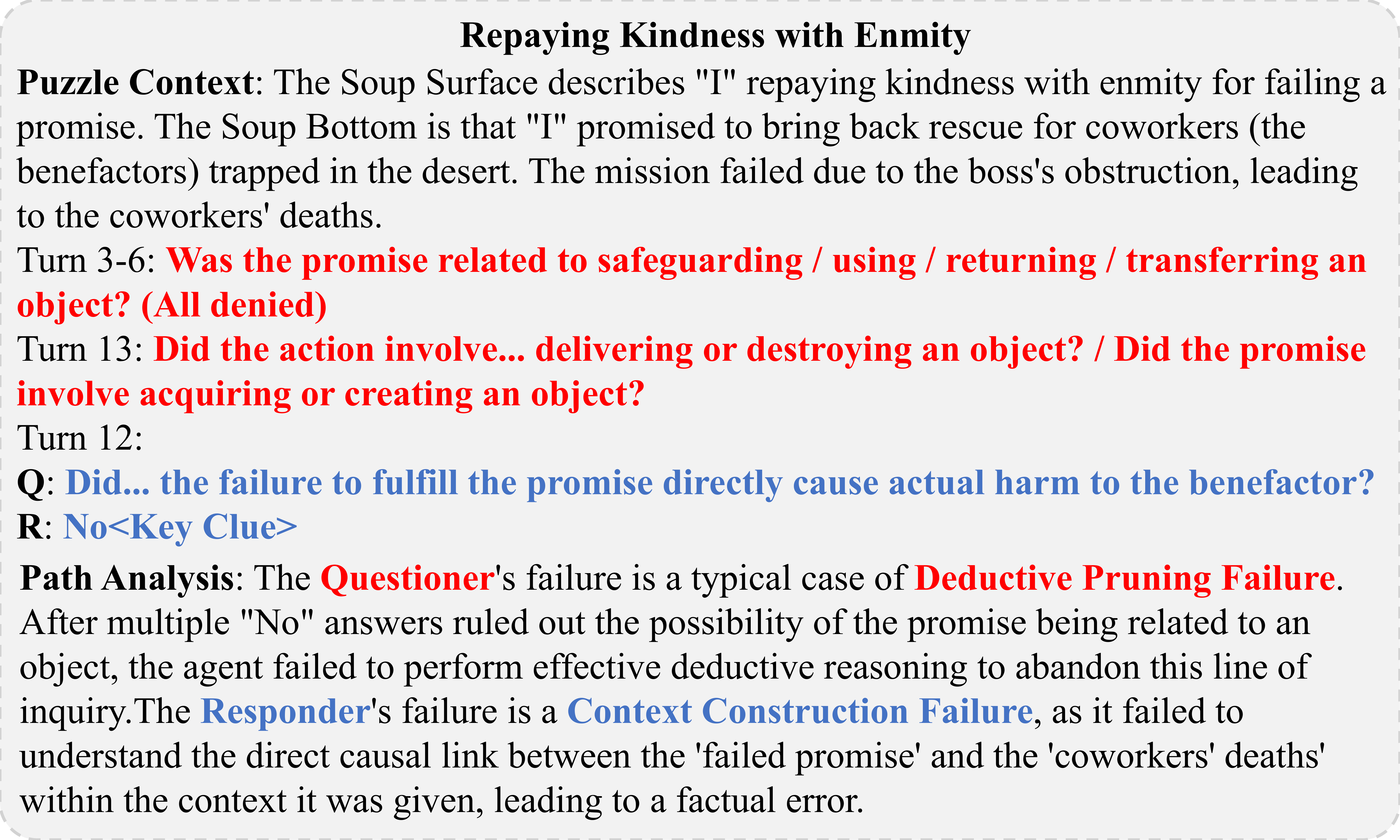}
    \caption{Case Studies of Four Typical Failure Paradigms}
    \label{fig:analysis}
\end{figure}

Our qualitative analysis categorizes the failures of LLM in imaginative reasoning into four distinct levels, progressing from the micro to the macro. Figure~\ref{fig:analysis}, through a specific case analysis, demonstrates two of these modes.

The most foundational failure is \textbf{Semantic Fixation}, which occurs at the level of word meaning interpretation. This stems from the model's reliance on the statistical inertia of its training data, causing it to rigidly default to a word's most high-frequency, literal meaning while ignoring contradictory contextual clues. This causes the entire reasoning process to start from a flawed premise.

This micro-level error then precipitates a more macroscopic \textbf{Context Construction Failure} at the scene level. The core deficit here is in integration and updating. Even when the model understands all individual clues, it fails to effectively splice these fragmented and sometimes contradictory pieces of information into a coherent global context. Then the imaginative process stalls 

\textbf{Logic Blind Spots} shows a higher-order bottleneck, a failure to reason about ``why''. Even with a correct factual context, the model struggles to conceive of the atypical causality that drives the scenario. Its reasoning paths are constrained by the common patterns within its training data, preventing it from proactively generating truly out-of-distribution hypotheses. This sharply defines the boundary of current LLM imagination: it excels at high-fidelity inference within its experiential space but lacks the creative leap required for genuinely novel solutions.

Finally, \textbf{Deductive Pruning Failure} is a fundamental deficit at the process and methodology level. This is less about content comprehension and more about the rigor of the reasoning process itself. The model ineffectively uses negative feedback to systematically eliminate falsified branches of the possibility space. Instead, it pursues redundant exploration down disproven paths. This demonstrates that model lacks the ability to adjust its own line of inquiry which renders its imaginative process both disordered and inefficient.

\subsection{Ablation study}
To verify the necessity of each agent in our cognitive architecture, we conducted ablation studies by selectively removing key components. All variants use deepseek-r1 as the base model and were tested on the original dataset from \textit{TurtleSoup-Bench}. The results are presented in Table~\ref{tab:ablation_deepseek}.

\begin{table}[h]
\renewcommand{\arraystretch}{0.9}
\centering
\setlength{\tabcolsep}{3pt} 
\begin{tabularx}{\columnwidth}{@{}Xcccc@{}}
\toprule
\textbf{Agent Config} & \textbf{$S_{\text{logic}}$} & \textbf{$S_{\text{details}}$} & \textbf{$S_{\text{conclusion}}$} & \textbf{$S_{\text{overall}}$} \\
\midrule
Mosaic-Agent     & 54.72 & 56.68 & 59.45 & 57.14 \\
w/o Deliberation    & 55.91 & 54.10 & 47.10 & 51.76 \\
w/o Meta-Cog.       & 56.14 & 50.51 & 47.70 & 51.07 \\
w/o Pruning         & 51.86 & 52.94 & 45.95 & 49.80 \\
w/o Key Clue        & 46.84 & 45.85 & 47.45 & 46.73 \\
w/o All agents     & 46.93 & 51.91 & 41.55 & 46.24 \\
\bottomrule
\end{tabularx}
\caption{Ablation study of the Mosaic-Agent framework driven by the deepseek-r1 model, evaluated on the original dataset from \textit{TurtleSoup-Bench}.}
\label{tab:ablation_deepseek}
\end{table}

\textbf{Deliberation agent.}
Removing the Deliberation Agent significantly harms performance (57.14 → 51.76), as the agent loses the ability to integrate scattered clues into a coherent narrative over long interactions. Its reasoning chain collapses, highlighting that a simple generative model is insufficient for such complex tasks.

\textbf{Meta-cognition agent.}
Disabling the Meta-cognition Agent (57.14 → 51.07) removes the agent's ability to adapt its strategy to the puzzle's genre. Without it, the agent applies a single, generic method to all problems, failing to target the core of the issue. This proves that advanced strategic planning requires explicit architectural support.

\textbf{Optimal Pruning}
Removing the Optimal Selection step (57.14 → 49.80) forces the agent to use its first, probabilistically generated idea instead of screening multiple candidates. This bypass of self-correction leads to less optimal decisions with more logical flaws, confirming that treating LLM outputs as proposals to be evaluated is key to reliability.

\textbf{Key Clue Mechanism}
Without the Key Clue mechanism, where the Responder no longer provides the \texttt{<Key Clue>} signal, causes the most severe performance collapse (57.14 → 46.73). This tag acts as a high signal-to-noise supervisory signal; without it, the agent's exploration degrades into a near-random walk amid low-value feedback. This finding suggests the quality of environmental feedback is a primary driver of exploration efficiency.

\textbf{Synergistic Effect of the Architecture}
Finally, the baseline agent (w/o All agents) scored 46.24, nearly identical to the score when only the Key Clue mechanism was removed (46.73). This reveals a crucial architectural synergy: sophisticated modules like Deliberation and Meta-cognition are ineffective without the high-quality information stream provided by the Key Clue mechanism. The ability to identify high-value information is thus a prerequisite for higher-order reasoning.

\section{Conclusion}
In this paper, we introduce a complete framework to probe the imaginative reasoning of LLMs, a critical yet under-explored capability for agents in information-scarce environments. Our framework consists of three core components: \textit{TurtleSoup-Bench}, the first large-scale, interactive benchmark for this task; Mosaic-Agent, an agent with a novel deliberative cognitive architecture; and a unique multi-dimensional evaluation protocol. Unlike prior work that assesses static outcomes or social dynamics, our research pioneers the evaluation of the exploratory reasoning process itself. Experiments conducted on our benchmark reveal the limitations of current state-of-the-art LLMs and validate our approach for analyzing the quality of an agent's reasoning path. This work establishes a new standard for evaluating imaginative reasoning and encourages a shift in focus from final outcomes to the dynamic process of inquiry.
\bibliography{anonymous-submission-latex-2026}

\begin{thebibliography}{46}
\providecommand{\natexlab}[1]{#1}

\bibitem[{Anthropic(2025)}]{anthropic2025}
Anthropic, C. 2025.
\newblock 3.7 sonnet and claude code.

\bibitem[{Bai et~al.(2024)Bai, Liu, Bu, He, Liu, Zhou, Lin, Su, Ge, Zheng, and Ouyang}]{bai-etal-2024-mt}
Bai, G.; Liu, J.; Bu, X.; He, Y.; Liu, J.; Zhou, Z.; Lin, Z.; Su, W.; Ge, T.; Zheng, B.; and Ouyang, W. 2024.
\newblock {MT}-Bench-101: A Fine-Grained Benchmark for Evaluating Large Language Models in Multi-Turn Dialogues.
\newblock In Ku, L.-W.; Martins, A.; and Srikumar, V., eds., \emph{Proceedings of the 62nd Annual Meeting of the Association for Computational Linguistics (Volume 1: Long Papers)}, 7421--7454. Bangkok, Thailand: Association for Computational Linguistics.

\bibitem[{Binz and Schulz(2023)}]{MarcelPANS}
Binz, M.; and Schulz, E. 2023.
\newblock Using cognitive psychology to understand GPT-3.
\newblock \emph{Proceedings of the National Academy of Sciences}, 120(6): e2218523120.

\bibitem[{Bubeck et~al.(2023)Bubeck, Chadrasekaran, Eldan, Gehrke, Horvitz, Kamar, Lee, Lee, Li, Lundberg et~al.}]{bubeck2023sparks}
Bubeck, S.; Chadrasekaran, V.; Eldan, R.; Gehrke, J.; Horvitz, E.; Kamar, E.; Lee, P.; Lee, Y.~T.; Li, Y.; Lundberg, S.; et~al. 2023.
\newblock Sparks of artificial general intelligence: Early experiments with gpt-4.

\bibitem[{Cai et~al.(2025)Cai, Gu, Du, Ye, Cao, Xu, Feng, and Chen}]{cai2025}
Cai, Y.; Gu, Z.; Du, Z.; Ye, Z.; Cao, S.; Xu, Y.; Feng, H.; and Chen, P. 2025.
\newblock MIRAGE: Exploring How Large Language Models Perform in Complex Social Interactive Environments.
\newblock arXiv:2501.01652.

\bibitem[{Comanici et~al.(2025)Comanici, Bieber, Schaekermann, Pasupat, Sachdeva, Dhillon, Blistein, Ram, Zhang, Rosen et~al.}]{comanici2025gemini}
Comanici, G.; Bieber, E.; Schaekermann, M.; Pasupat, I.; Sachdeva, N.; Dhillon, I.; Blistein, M.; Ram, O.; Zhang, D.; Rosen, E.; et~al. 2025.
\newblock Gemini 2.5: Pushing the frontier with advanced reasoning, multimodality, long context, and next generation agentic capabilities.

\bibitem[{Curvo(2025)}]{curvo2025}
Curvo, P. M.~P. 2025.
\newblock The Traitors: Deception and Trust in Multi-Agent Language Model Simulations.
\newblock arXiv:2505.12923.

\bibitem[{Dubey et~al.(2024)Dubey, Jauhri, Pandey, Kadian, Al-Dahle, Letman, Mathur, Schelten, Yang, Fan et~al.}]{dubey2024llama}
Dubey, A.; Jauhri, A.; Pandey, A.; Kadian, A.; Al-Dahle, A.; Letman, A.; Mathur, A.; Schelten, A.; Yang, A.; Fan, A.; et~al. 2024.
\newblock The llama 3 herd of models.

\bibitem[{Eriksson et~al.(2025)Eriksson, Purificato, Noroozian, Vinagre, Chaslot, Gomez, and Fernandez-Llorca}]{eriksson2025}
Eriksson, M.; Purificato, E.; Noroozian, A.; Vinagre, J.; Chaslot, G.; Gomez, E.; and Fernandez-Llorca, D. 2025.
\newblock Can We Trust AI Benchmarks? An Interdisciplinary Review of Current Issues in AI Evaluation.
\newblock arXiv:2502.06559.

\bibitem[{Guo et~al.(2025)Guo, Yang, Zhang, Song, Zhang, Xu, Zhu, Ma, Wang, Bi et~al.}]{guo2025deepseek}
Guo, D.; Yang, D.; Zhang, H.; Song, J.; Zhang, R.; Xu, R.; Zhu, Q.; Ma, S.; Wang, P.; Bi, X.; et~al. 2025.
\newblock Deepseek-r1: Incentivizing reasoning capability in llms via reinforcement learning.

\bibitem[{Hendrycks et~al.(2021)Hendrycks, Burns, Basart, Zou, Mazeika, Song, and Steinhardt}]{hendrycks2021measuring}
Hendrycks, D.; Burns, C.; Basart, S.; Zou, A.; Mazeika, M.; Song, D.; and Steinhardt, J. 2021.
\newblock Measuring Massive Multitask Language Understanding.
\newblock In \emph{International Conference on Learning Representations}.

\bibitem[{Hong et~al.(2024)Hong, Zhuge, Chen, Zheng, Cheng, Wang, Zhang, Wang, Yau, Lin, Zhou, Ran, Xiao, Wu, and Schmidhuber}]{hong2024metagpt}
Hong, S.; Zhuge, M.; Chen, J.; Zheng, X.; Cheng, Y.; Wang, J.; Zhang, C.; Wang, Z.; Yau, S. K.~S.; Lin, Z.; Zhou, L.; Ran, C.; Xiao, L.; Wu, C.; and Schmidhuber, J. 2024.
\newblock Meta{GPT}: Meta Programming for A Multi-Agent Collaborative Framework.
\newblock In \emph{The Twelfth International Conference on Learning Representations}.

\bibitem[{Hsia et~al.(2024)Hsia, Pruthi, Singh, and Lipton}]{hsia-etal-2024}
Hsia, J.; Pruthi, D.; Singh, A.; and Lipton, Z. 2024.
\newblock Goodhart{'}s Law Applies to {NLP}{'}s Explanation Benchmarks.
\newblock In Graham, Y.; and Purver, M., eds., \emph{Findings of the Association for Computational Linguistics: EACL 2024}, 1322--1335. St. Julian{'}s, Malta: Association for Computational Linguistics.

\bibitem[{Hurst et~al.(2024)Hurst, Lerer, Goucher, Perelman, Ramesh, Clark, Ostrow, Welihinda, Hayes, Radford et~al.}]{hurst2024gpt}
Hurst, A.; Lerer, A.; Goucher, A.~P.; Perelman, A.; Ramesh, A.; Clark, A.; Ostrow, A.; Welihinda, A.; Hayes, A.; Radford, A.; et~al. 2024.
\newblock Gpt-4o system card.

\bibitem[{Kidd and Hayden(2015)}]{Kidd2015-ze}
Kidd, C.; and Hayden, B.~Y. 2015.
\newblock The psychology and neuroscience of curiosity.
\newblock \emph{Neuron}, 88(3): 449--460.

\bibitem[{Lan et~al.(2024)Lan, Hu, Wang, Wang, Ye, Zhao, Lim, Xiong, and Wang}]{lan-etal-2024}
Lan, Y.; Hu, Z.; Wang, L.; Wang, Y.; Ye, D.; Zhao, P.; Lim, E.-P.; Xiong, H.; and Wang, H. 2024.
\newblock {LLM}-Based Agent Society Investigation: Collaboration and Confrontation in Avalon Gameplay.
\newblock In Al-Onaizan, Y.; Bansal, M.; and Chen, Y.-N., eds., \emph{Proceedings of the 2024 Conference on Empirical Methods in Natural Language Processing}, 128--145. Miami, Florida, USA: Association for Computational Linguistics.

\bibitem[{Li et~al.(2025)Li, Bao, Ma, Li, Wang, Men, Zhang, Feng, Liu, and Lin}]{li2025}
Li, X.; Bao, K.; Ma, Y.; Li, M.; Wang, W.; Men, R.; Zhang, Y.; Feng, F.; Liu, D.; and Lin, J. 2025.
\newblock MTR-Bench: A Comprehensive Benchmark for Multi-Turn Reasoning Evaluation.
\newblock arXiv:2505.17123.

\bibitem[{Liang et~al.(2023)Liang, He, tse Huang, Wang, Jiao, Wang, Yang, Tu, Shi, and Wang}]{liang2023}
Liang, T.; He, Z.; tse Huang, J.; Wang, W.; Jiao, W.; Wang, R.; Yang, Y.; Tu, Z.; Shi, S.; and Wang, X. 2023.
\newblock Leveraging Word Guessing Games to Assess the Intelligence of Large Language Models.
\newblock arXiv:2310.20499.

\bibitem[{Liu et~al.(2023)Liu, Iter, Xu, Wang, Xu, and Zhu}]{liu-etal-2023-g}
Liu, Y.; Iter, D.; Xu, Y.; Wang, S.; Xu, R.; and Zhu, C. 2023.
\newblock {G}-Eval: {NLG} Evaluation using Gpt-4 with Better Human Alignment.
\newblock In Bouamor, H.; Pino, J.; and Bali, K., eds., \emph{Proceedings of the 2023 Conference on Empirical Methods in Natural Language Processing}, 2511--2522. Singapore: Association for Computational Linguistics.

\bibitem[{Ma et~al.(2024)Ma, Zhang, Zhu, Yang, Yang, Jin, Lan, Kong, and He}]{Ma2024}
Ma, C.; Zhang, J.; Zhu, Z.; Yang, C.; Yang, Y.; Jin, Y.; Lan, Z.; Kong, L.; and He, J. 2024.
\newblock AgentBoard: An Analytical Evaluation Board of Multi-turn LLM Agents.
\newblock In Globerson, A.; Mackey, L.; Belgrave, D.; Fan, A.; Paquet, U.; Tomczak, J.; and Zhang, C., eds., \emph{Advances in Neural Information Processing Systems}, volume~37, 74325--74362. Curran Associates, Inc.

\bibitem[{Park et~al.(2023)Park, O'Brien, Cai, Morris, Liang, and Bernstein}]{park2023generative}
Park, J.~S.; O'Brien, J.; Cai, C.~J.; Morris, M.~R.; Liang, P.; and Bernstein, M.~S. 2023.
\newblock Generative Agents: Interactive Simulacra of Human Behavior.
\newblock In \emph{Proceedings of the 36th Annual ACM Symposium on User Interface Software and Technology}, UIST '23. New York, NY, USA: Association for Computing Machinery.
\newblock ISBN 9798400701320.

\bibitem[{Press et~al.(2023)Press, Zhang, Min, Schmidt, Smith, and Lewis}]{press-etal-2023-measuring}
Press, O.; Zhang, M.; Min, S.; Schmidt, L.; Smith, N.; and Lewis, M. 2023.
\newblock Measuring and Narrowing the Compositionality Gap in Language Models.
\newblock In Bouamor, H.; Pino, J.; and Bali, K., eds., \emph{Findings of the Association for Computational Linguistics: EMNLP 2023}, 5687--5711. Singapore: Association for Computational Linguistics.

\bibitem[{Sato, Ozaki, and Yokoyama(2024)}]{sato-etal-2024}
Sato, T.; Ozaki, S.; and Yokoyama, D. 2024.
\newblock An Implementation of Werewolf Agent That does not Truly Trust {LLM}s.
\newblock In Kano, Y., ed., \emph{Proceedings of the 2nd International AIWolfDial Workshop}, 58--67. Tokyo, Japan: Association for Computational Linguistics.

\bibitem[{Schacter, Addis, and Buckner(2007)}]{Schacter2007}
Schacter, D.~L.; Addis, D.~R.; and Buckner, R.~L. 2007.
\newblock Remembering the past to imagine the future: the prospective brain.
\newblock \emph{Nature Reviews Neuroscience}, 8(9): 657--661.

\bibitem[{Shinn et~al.(2023)Shinn, Cassano, Gopinath, Narasimhan, and Yao}]{shinn2023reflexion}
Shinn, N.; Cassano, F.; Gopinath, A.; Narasimhan, K.~R.; and Yao, S. 2023.
\newblock Reflexion: language agents with verbal reinforcement learning.
\newblock In \emph{Thirty-seventh Conference on Neural Information Processing Systems}.

\bibitem[{Tang(2025)}]{TangWebsite}
Tang, H. 2025.
\newblock Henre Tang's Personal Website.
\newblock \url{https://tanghenre.com/}.
\newblock Accessed: 2025-07-28.

\bibitem[{Wang et~al.(2024{\natexlab{a}})Wang, Xie, Jiang, Mandlekar, Xiao, Zhu, Fan, and Anandkumar}]{wang2024voyager}
Wang, G.; Xie, Y.; Jiang, Y.; Mandlekar, A.; Xiao, C.; Zhu, Y.; Fan, L.; and Anandkumar, A. 2024{\natexlab{a}}.
\newblock Voyager: An Open-Ended Embodied Agent with Large Language Models.
\newblock \emph{Transactions on Machine Learning Research}.

\bibitem[{Wang et~al.(2025)Wang, Lian, Huang, Dai, Li, Chen, Xie, and Wen}]{wang-etal-2025}
Wang, L.; Lian, J.; Huang, Y.; Dai, Y.; Li, H.; Chen, X.; Xie, X.; and Wen, J.-R. 2025.
\newblock {C}haracter{B}ox: Evaluating the Role-Playing Capabilities of {LLM}s in Text-Based Virtual Worlds.
\newblock In Chiruzzo, L.; Ritter, A.; and Wang, L., eds., \emph{Proceedings of the 2025 Conference of the Nations of the Americas Chapter of the Association for Computational Linguistics: Human Language Technologies (Volume 1: Long Papers)}, 6372--6391. Albuquerque, New Mexico: Association for Computational Linguistics.
\newblock ISBN 979-8-89176-189-6.

\bibitem[{Wang et~al.(2024{\natexlab{b}})Wang, Peng, Que, Liu, Zhou, Wu, Guo, Gan, Ni, Yang, Zhang, Zhang, Ouyang, Xu, Huang, Fu, and Peng}]{wang-etal-2024}
Wang, N.; Peng, Z.; Que, H.; Liu, J.; Zhou, W.; Wu, Y.; Guo, H.; Gan, R.; Ni, Z.; Yang, J.; Zhang, M.; Zhang, Z.; Ouyang, W.; Xu, K.; Huang, W.; Fu, J.; and Peng, J. 2024{\natexlab{b}}.
\newblock {R}ole{LLM}: Benchmarking, Eliciting, and Enhancing Role-Playing Abilities of Large Language Models.
\newblock In Ku, L.-W.; Martins, A.; and Srikumar, V., eds., \emph{Findings of the Association for Computational Linguistics: ACL 2024}, 14743--14777. Bangkok, Thailand: Association for Computational Linguistics.

\bibitem[{Wei, Chen, and Xu(2025)}]{wei2025}
Wei, C.; Chen, J.; and Xu, J. 2025.
\newblock Exploring Large Language Models for Word Games:Who is the Spy?
\newblock arXiv:2503.15235.

\bibitem[{{Wikipedia contributors}(2024)}]{wiki:SituationPuzzle}
{Wikipedia contributors}. 2024.
\newblock Situation puzzle --- {Wikipedia}{,} The Free Encyclopedia.
\newblock [Online; accessed 25-July-2024].

\bibitem[{Wu et~al.(2024)Wu, Shi, Sun, and Liu}]{wu-etal-2024}
Wu, D.; Shi, H.; Sun, Z.; and Liu, B. 2024.
\newblock Deciphering Digital Detectives: Understanding {LLM} Behaviors and Capabilities in Multi-Agent Mystery Games.
\newblock In Ku, L.-W.; Martins, A.; and Srikumar, V., eds., \emph{Findings of the Association for Computational Linguistics: ACL 2024}, 8225--8291. Bangkok, Thailand: Association for Computational Linguistics.

\bibitem[{Wu et~al.(2023)Wu, Khasahmadi, Katz, Jayaraman, Pu, Willis, and Liu}]{wu2023cadllm_workshop}
Wu, S.; Khasahmadi, A.; Katz, M.; Jayaraman, P.~K.; Pu, Y.; Willis, K.; and Liu, B. 2023.
\newblock {Cad-llm: Large language model for cad generation}.
\newblock In \emph{NeurIPS 2023 Workshop on Machine Learning for Creativity and Design}.

\bibitem[{Wu et~al.(2025{\natexlab{a}})Wu, Khasahmadi, Katz, Jayaraman, Pu, Willis, and Liu}]{Wu2025eccv}
Wu, S.; Khasahmadi, A.~H.; Katz, M.; Jayaraman, P.~K.; Pu, Y.; Willis, K.; and Liu, B. 2025{\natexlab{a}}.
\newblock CadVLM: Bridging Language and Vision in the Generation of Parametric CAD Sketches.
\newblock In Leonardis, A.; Ricci, E.; Roth, S.; Russakovsky, O.; Sattler, T.; and Varol, G., eds., \emph{Computer Vision -- ECCV 2024}, 368--384. Cham: Springer Nature Switzerland.

\bibitem[{Wu et~al.(2025{\natexlab{b}})Wu, Zhang, Li, Effaty, Ataei, and Liu}]{wu2025seeing}
Wu, S.; Zhang, H.; Li, Y.; Effaty, F.; Ataei, A.; and Liu, B. 2025{\natexlab{b}}.
\newblock Seeing Beyond Words: MatVQA for Challenging Visual-Scientific Reasoning in Materials Science.
\newblock arXiv:2505.18319.

\bibitem[{Xi et~al.(2025)Xi, Chen, Guo et~al.}]{xi2023rise}
Xi, Z.; Chen, W.; Guo, X.; et~al. 2025.
\newblock The rise and potential of large language model based agents: a survey.
\newblock \emph{Science China Information Sciences}, 68: 121101.

\bibitem[{Xu et~al.(2025)Xu, Gu, Yu, Wu, and Wang}]{xu2025learning}
Xu, Z.; Gu, W.; Yu, C.; Wu, Y.; and Wang, Y. 2025.
\newblock Learning Strategic Language Agents in the Werewolf Game with Iterative Latent Space Policy Optimization.
\newblock In \emph{Forty-second International Conference on Machine Learning}.

\bibitem[{Xu et~al.(2024)Xu, Yu, Fang, Wang, and Wu}]{xu2024}
Xu, Z.; Yu, C.; Fang, F.; Wang, Y.; and Wu, Y. 2024.
\newblock Language agents with reinforcement learning for strategic play in the Werewolf game.
\newblock In \emph{Proceedings of the 41st International Conference on Machine Learning}, ICML'24. JMLR.org.

\bibitem[{Yang et~al.(2025{\natexlab{a}})Yang, Li, Yang, Zhang, Hui, Zheng, Yu, Gao, Huang, Lv et~al.}]{yang2025qwen3}
Yang, A.; Li, A.; Yang, B.; Zhang, B.; Hui, B.; Zheng, B.; Yu, B.; Gao, C.; Huang, C.; Lv, C.; et~al. 2025{\natexlab{a}}.
\newblock Qwen3 technical report.

\bibitem[{Yang and Jin(2024)}]{yang2024makesgoodstorymeasure}
Yang, D.; and Jin, Q. 2024.
\newblock What Makes a Good Story and How Can We Measure It? A Comprehensive Survey of Story Evaluation.
\newblock arXiv:2408.14622.

\bibitem[{Yang et~al.(2025{\natexlab{b}})Yang, Chen, Zhang, Zhao, Qian, Wang, Wang, Koripella, Movahedi, Li, Ji, Zhang, and Zhang}]{yang2025embodiedbench}
Yang, R.; Chen, H.; Zhang, J.; Zhao, M.; Qian, C.; Wang, K.; Wang, Q.; Koripella, T.~V.; Movahedi, M.; Li, M.; Ji, H.; Zhang, H.; and Zhang, T. 2025{\natexlab{b}}.
\newblock EmbodiedBench: Comprehensive Benchmarking Multi-modal Large Language Models for Vision-Driven Embodied Agents.
\newblock In \emph{Forty-second International Conference on Machine Learning}.

\bibitem[{Yao et~al.(2023)Yao, Zhao, Yu, Du, Shafran, Narasimhan, and Cao}]{yao2023react}
Yao, S.; Zhao, J.; Yu, D.; Du, N.; Shafran, I.; Narasimhan, K.~R.; and Cao, Y. 2023.
\newblock ReAct: Synergizing Reasoning and Acting in Language Models.
\newblock In \emph{The Eleventh International Conference on Learning Representations}.

\bibitem[{Yu et~al.(2025)Yu, Shen, Meng, Lee, Yin, Cui, Xu, Zhu, Shi, Li, and Smola}]{yu2025}
Yu, P.; Shen, D.; Meng, S.; Lee, J.; Yin, W.; Cui, A.~Y.; Xu, Z.; Zhu, Y.; Shi, X.; Li, M.; and Smola, A. 2025.
\newblock RPGBENCH: Evaluating Large Language Models as Role-Playing Game Engines.
\newblock arXiv:2502.00595.

\bibitem[{Yu et~al.(2024)Yu, Song, Fang, Shi, Zheng, Wang, Niu, and Li}]{yu2024turtlebench}
Yu, Q.; Song, S.; Fang, K.; Shi, Y.; Zheng, Z.; Wang, H.; Niu, S.; and Li, Z. 2024.
\newblock TurtleBench: Evaluating Top Language Models via Real-World Yes/No Puzzles.
\newblock arXiv:2410.05262.

\bibitem[{Zhang et~al.(2025)Zhang, Lan, Chen, Wang, Wang, and Wang}]{zhang2025dvm}
Zhang, Z.; Lan, Y.; Chen, Y.; Wang, L.; Wang, X.; and Wang, H. 2025.
\newblock DVM: Towards Controllable LLM Agents in Social Deduction Games.
\newblock In \emph{ICASSP 2025-2025 IEEE International Conference on Acoustics, Speech and Signal Processing (ICASSP)}. IEEE.

\bibitem[{Zhu et~al.(2024)Zhu, Zhao, Du, Gui, and He}]{Qinglin2024}
Zhu, Q.; Zhao, R.; Du, J.; Gui, L.; and He, Y. 2024.
\newblock PLAYER*: Enhancing LLM-based Multi-Agent Communication and Interaction in Murder Mystery Games.
\newblock \emph{CoRR}, abs/2404.17662.

\end{thebibliography}

\clearpage 
\section*{Appendix} 
\renewcommand{\thesection}{\Alph{section}} 
\setcounter{section}{0} 

\section{TurtleSoup-Bench Benchmark and Game Environment}
\label{app:benchmark}

This section provides a brief overview of the ``Turtle Soup" game and presents several representative examples from our \textit{TurtleSoup-Bench} to illustrate the diversity and complexity of the tasks. Further details on our implementation and benchmark will be made available in the future.

\subsection{Game Introduction}
The "Turtle Soup" game, also known as a Situation Puzzle, is a collaborative reasoning game played between two roles: a storyteller and one or more players. The storyteller, acting as an omniscient "God," knows the complete, often unusual, story behind an event. They begin by presenting only the final, puzzling outcome of this story to the players. The players' objective is to uncover the full narrative by asking a series of questions. The storyteller is strictly limited to answering with only ``Yes,'' ``No'', ``Unknown'', or ``Irrelevant''.

\subsection{Several Examples}
Our \textit{TurtleSoup-Bench} encompasses a wide range of narrative genres. For the purpose of this appendix and to ensure a comfortable reading experience for all audiences, we have consciously omitted examples from categories that may contain graphic or potentially disturbing content, such as ``Crime Thriller'' and ``Mind Game''. This section, therefore, presents one representative puzzle from each of the remaining non-sensitive genres. 

\noindent\textbf{Supernatural Fantasy: The Slide}

\noindent\textbf{Soup Surface:}
\par\textit{Near programmer Wang's home, there is a park. Every day when Wang passed the park on his way home from working overtime, he would see a child in a plaid shirt standing by the slide with a terrified look. One day, Wang didn't work overtime and, on a whim, went down the slide. After sliding down, Wang was horrified. What did he see?}

\noindent\textbf{Soup Bottom:}
\par\textit{It turned out Wang saw himself on his way home from work. The slide was a time machine. When Wang slid down, he turned into the child. The child's plaid shirt was actually programmer Wang's. The terrified child he saw every time was himself after sliding down.}

\noindent\textbf{Key Clue Library:}
\begin{itemize}
    \item \textit{The slide has an unnatural function.}
    \item \textit{The slide is related to time.}
    \item \textit{The terrified child seen every day is the same person.}
    \item \textit{The child's plaid shirt is related to Wang.}
    \item \textit{After sliding down, Wang saw himself.}
    \item \textit{Sliding down caused a change in age.}
    \item \textit{The child was the younger version of Wang himself.}
\end{itemize}

\noindent\textbf{Constant Change: Old Man}
\label{ex: Cons}

\noindent\textbf{Soup Surface:}
\par\textit{The old man ferried the man across the river. The man crossed the river, but the old man died in the river.}

\noindent\textbf{Soup Bottom:}
\par\textit{The man was a chivalrous thief who was wanted for killing a tyrannical king. The old man recognized him when he fled to the riverbank and took him across the river. He repeatedly urged the old man not to reveal his whereabouts. The old man then committed suicide to prove to him that he would never leak the secret.}

\noindent\textbf{Key Clue Library:}
\begin{itemize}
    \item \textit{The man's identity was special; he was evading pursuit.}
    \item \textit{The old man knew the man's identity.}
    \item \textit{Before and after crossing the river, the man emphasized the importance of secrecy to the old man.}
    \item \textit{The old man's death was not an accident but his own choice.}
    \item \textit{The old man chose death to protect the man's secret.}
\end{itemize}

\noindent\textbf{Clever Logic: Handgun}
\label{ex:logic}

\noindent\textbf{Soup Surface:}
\par\textit{He returned home and saw a woman holding a handgun, saying she was going to kill him.}

\noindent\textbf{Soup Bottom:}
\par\textit{He is a drama critic who once reviewed a male actor playing a murderer as having poor acting skills. This actor then disguised himself as a woman wanting to kill him and hid in his house. If he didn't realize the actor didn't actually want to kill him, it would prove the actor's acting skills.}

\noindent\textbf{Key Clue Library:}
\begin{itemize}
    \item \textit{The gun wielder's threat was to achieve another purpose, not genuinely to kill.}
    \item \textit{The gun wielder's female appearance was a disguise.}
    \item \textit{The man's past professional actions were the direct cause of this incident.}
    \item \textit{This scene was a carefully arranged performance, not a real attack.}
    \item \textit{The gun wielder's purpose was to prove a certain professional skill to the man.}
\end{itemize}

\noindent\subsubsection{Original Data: Box}
\label{ex:logic}

\noindent\textbf{Soup Surface:}
\par\textit{While tidying up my belongings, I found a box containing a tattered photo, and tears streamed down my face.}

\noindent\textbf{Soup Bottom:}
\par\textit{When I was little, I was mischievous and accidentally set fire to my room with a match. I was so scared that I ran away without telling my parents, who were at home, taking only a family photo that they had kept on me in case I got lost. My family died in that fire, and I was forced to leave my hometown. For many years, I have lived in regret. Years later, I saw this photo again. The house in the photo was my former home, with my parents standing on either side, smiling and holding my hands. This stirred up years of unresolved guilt and remorse deep within me, causing tears to stream down my face.}

\noindent\textbf{Key Clue Library:}
\begin{itemize}
    \item \textit{I once caused a fire.}
    \item \textit{The fire led to the death of my family.}
    \item \textit{I fled the fire scene out of fear.}
    \item \textit{The only item I took was a family photo.}
    \item \textit{The photo evoked memories of past guilt and remorse.}
\end{itemize}

\section{Full Interaction and Case Study}
\label{app:case_study}

This section presents the complete, unabridged, turn-by-turn interaction log of the Mosaic-Agent solving the puzzle ``The Last Rose.'' It demonstrates the agent's reasoning process in action, culminating in the generation of its final Analysis and Proposal Set (APS) before concluding the puzzle.

\subsection{Interaction Log for ``The Last Rose''}

\par\noindent\textbf{Turn 1}
\par\noindent\textbf{Questioner:} Was the ``Sorry'' on the card related to a specific event?
\par\noindent\textbf{Responder:} Yes \texttt{<Key Clue>}
\par

\par\noindent\textbf{Turn 2}
\par\noindent\textbf{Questioner:} Was the sender someone the narrator knows?
\par\noindent\textbf{Responder:} Yes
\par

\par\noindent\textbf{Turn 3}
\par\noindent\textbf{Questioner:} Is the sender the narrator's lover?
\par\noindent\textbf{Responder:} Yes \texttt{<Key Clue>}
\par

\par\noindent\textbf{Turn 4}
\par\noindent\textbf{Questioner:} Did the specific event involve a quarrel between the lovers?
\par\noindent\textbf{Responder:} Yes \texttt{<Key Clue>}
\par

\par\noindent\textbf{Turn 5}
\par\noindent\textbf{Questioner:} After ignoring the card, did the lover leave the narrator?
\par\noindent\textbf{Responder:} Yes \texttt{<Key Clue>}
\par

\par\noindent\textbf{Turn 6}
\par\noindent\textbf{Questioner:} Was the regret ``a few days later'' triggered by finding a specific item left by the lover?
\par\noindent\textbf{Responder:} Unknown \texttt{<Key Clue>}
\par

\par\noindent\textbf{Turn 7}
\par\noindent\textbf{Questioner:} Was the regret triggered by the narrator's own recollection of the quarrel?
\par\noindent\textbf{Responder:} No
\par

\par\noindent\textbf{Turn 8}
\par\noindent\textbf{Questioner:} At the time of regret, had the narrator learned new information about the lover from a third party?
\par\noindent\textbf{Responder:} Unknown
\par

\par\noindent\textbf{Turn 9}
\par\noindent\textbf{Questioner:} Was the regret triggered by the narrator learning that the lover had an accident?
\par\noindent\textbf{Responder:} Yes \texttt{<Key Clue>}
\par

\par\noindent\textbf{Turn 10}
\par\noindent\textbf{Questioner:} Did the accident directly cause the lover's death?
\par\noindent\textbf{Responder:} No
\par

\par\noindent\textbf{Turn 11}
\par\noindent\textbf{Questioner:} Is the lover still alive?
\par\noindent\textbf{Responder:} No \texttt{<Key Clue>}
\par

\par\noindent\textbf{Turn 12}
\par\noindent\textbf{Questioner:} Was the lover's death indirectly caused by the accident?
\par\noindent\textbf{Responder:} No
\par

\par\noindent\textbf{Turn 13}
\par\noindent\textbf{Questioner:} Was the lover's death caused by suicide?
\par\noindent\textbf{Responder:} No
\par

\par\noindent\textbf{Turn 14}
\par\noindent\textbf{Questioner:} Was the lover's death caused by an illness?
\par\noindent\textbf{Responder:} Yes \texttt{<Key Clue>}
\par

\par\noindent\textbf{Turn 15}
\par\noindent\textbf{Questioner:} Was the illness diagnosed before the flowers were sent?
\par\noindent\textbf{Responder:} Yes \texttt{<Key Clue>}
\par

\par\noindent\textbf{Turn 16}
\par\noindent\textbf{Questioner:} Was the accident directly related to the lover's illness (e.g., a worsening of the condition or death)?
\par\noindent\textbf{Responder:} Yes \texttt{<Key Clue>}
\par

\par\noindent\textbf{Turn 17}
\par\noindent\textbf{Questioner:} Did the accident's worsening of the illness accelerate the lover's death?
\par\noindent\textbf{Responder:} Unknown
\par

\par\noindent\textbf{Turn 18}
\par\noindent\textbf{Questioner:} Did the accident happen after the lover left the narrator?
\par\noindent\textbf{Responder:} Yes
\par

\par\noindent\textbf{Turn 19}
\par\noindent\textbf{Questioner:} Was the specific nature of the accident that the lover had a sudden health crisis and was sent to the hospital?
\par\noindent\textbf{Responder:} No
\par

\par\noindent\textbf{Turn 20}
\par\noindent\textbf{Questioner:} Did the accident happen at the lover's home after they left?
\par\noindent\textbf{Responder:} Unknown
\par

\par\noindent\textbf{Turn 21}
\par\noindent\textbf{Questioner:} Did the card only have ``Sorry'' written on it and no other information?
\par\noindent\textbf{Responder:} No \texttt{<Key Clue>}
\par

\par\noindent\textbf{Turn 22}
\par\noindent\textbf{Questioner:} Did the other information on the card directly relate to the lover's medical condition?
\par\noindent\textbf{Responder:} No
\par

\par\noindent\textbf{Turn 23}
\par\noindent\textbf{Questioner:} Did the other information on the card directly explain the quarrel?
\par\noindent\textbf{Responder:} No
\par

\par\noindent\textbf{Turn 24}
\par\noindent\textbf{Questioner:} Did the other information on the card contain a specific request to meet?
\par\noindent\textbf{Responder:} No
\par

\par\noindent\textbf{Turn 25}
\par\noindent\textbf{Questioner:} Did the other information on the card imply that the narrator should actively contact the lover?
\par\noindent\textbf{Responder:} Yes \texttt{<Key Clue>}
\par

\par\noindent\textbf{Turn 26}
\par\noindent\textbf{Questioner:} Did the other information on the card imply that the narrator should contact the lover, but the narrator failed to understand its meaning at the time?
\par\noindent\textbf{Responder:} Yes \texttt{<Key Clue>}
\par

\par\noindent\textbf{Turn 27}
\par\noindent\textbf{Questioner:} Was the reason the narrator failed to understand the card's implication due to their emotional state (e.g., anger)?
\par\noindent\textbf{Responder:} Yes \texttt{<Key Clue>}
\par

\par\noindent\textbf{Turn 28}
\par\noindent\textbf{Questioner:} Did the narrator's anger lead them to completely ignore the card's implied message?
\par\noindent\textbf{Responder:} Yes \texttt{<Key Clue>}
\par

\par\noindent\textbf{Turn 29}
\par\noindent\textbf{Questioner:} Was the specific trigger of the quarrel related to the lover's illness?
\par\noindent\textbf{Responder:} No
\par

\par\noindent\textbf{Turn 30}
\par\noindent\textbf{Questioner:} Was the specific trigger of the quarrel a misunderstanding of the lover by the narrator?
\par\noindent\textbf{Responder:} Unknown
\par

\subsection{Agent's Internal State at Turn 25}

To illustrate the agent's cognitive process, we present a snapshot of its internal state after Turn 25, a critical juncture in the puzzle. At this point, the agent has gathered most of the key clues and performs a global deliberation to first update its Belief State ($B_t$) and then generate a final Analysis and Proposal Set (APS) to guide its concluding actions.

\subsubsection{Belief State ($B_t$) Generation at Turn 25}

Based on the conversation history and all discovered key clues, the agent generates the following comprehensive summary of the situation, structured into details, abstract logic, and a narrative conclusion.

\par\vspace{0.5em}
\noindent\textbf{Details:}
\begin{itemize}
    \item The narrator received a bouquet of roses and a card.
    \item The card said ``Sorry'' but also contained other, implicit information.
    \item The sender was the narrator's lover.
    \item The apology was related to a specific quarrel between them.
    \item The narrator ignored the card and its message.
    \item After the card was ignored, the lover ``left,'' which means they died.
    \item The death was caused by a terminal illness.
    \item This illness had been diagnosed *before* the flowers were sent.
    \item The narrator's later regret was triggered by learning of the lover's death (referred to as an ``accident'').
    \item The implicit information on the card was a plea for the narrator to initiate contact.
\end{itemize}

\noindent\textbf{Logic:}
\begin{itemize}
    \item A relationship is strained by an initial conflict.
    \item One party, possessing critical uncommunicated information (a health crisis), sends a symbolic gift with a message attempting reconciliation.
    \item This message contains both an explicit apology and an implicit call to action.
    \item The receiving party, influenced by a negative emotional state from the initial conflict, fails to process the implicit call to action.
    \item Consequently, a final window for communication is irrevocably missed.
    \item The sending party succumbs to their pre-existing critical condition.
    \item Upon discovering the full context (the death and the missed final message), the receiving party experiences profound regret.
\end{itemize}

\noindent\textbf{Conclusion:}

The story stems from a quarrel between the narrator and their lover. The lover, who was secretly terminally ill, sent a bouquet of roses with a card as a final act of reconciliation and farewell. The card explicitly said ``Sorry'' and implicitly urged the narrator to make contact. However, consumed by anger from the quarrel, the narrator paid no attention to the card, thereby missing this final, urgent plea. A few days later, the lover passed away from their illness. When the narrator learned of the death, they were overwhelmed with regret, realizing they had not just ignored an apology but had unwittingly missed the last chance to say goodbye to their dying partner.

\subsubsection{Final Analysis and Proposal Set (APS) Generation}
Immediately following the generation of the Belief State above, the agent identifies remaining ambiguities to produce a final APS. This set guides the last few questions or the generation of the final summary.

\par\vspace{0.5em}
\noindent\textbf{Doubt 1:} What is the direct cause of the narrator's intense regret, beyond just the quarrel itself?
\par\noindent\textbf{Analysis 1:} The agent has confirmed the narrator was angry over a quarrel. The lover sent a card with an apology and a hidden message implying a need for contact. The narrator's anger caused them to completely ignore this hidden message. This missed signal is the core of the narrator's failure.
\par\noindent\textbf{Proposal 1:} The final summary must center on the narrator's emotional state causing them to miss a crucial, hidden message of reconciliation from their lover.

\par\vspace{0.5em}
\noindent\textbf{Doubt 2:} What transforms this missed opportunity into an irreversible tragedy?
\par\noindent\textbf{Analysis 2:} The agent confirmed the lover died from a pre-diagnosed terminal illness. The timing of the flowers and card now becomes clear: it was not just an apology, but a final goodbye. The narrator's inaction meant they missed their lover's final days and farewell.
\par\noindent\textbf{Proposal 2:} The conclusion must explicitly state that the lover was terminally ill and the flowers were a final act of love. The narrator's regret stems from realizing they missed the last chance to be with their dying partner, a fact they were ignorant of at the time.

\section{Detailed Failure Case Analysis}
\label{app:failure_cases}

\begin{figure*}[t!]
    \centering
    \includegraphics[width=\textwidth, page=1]{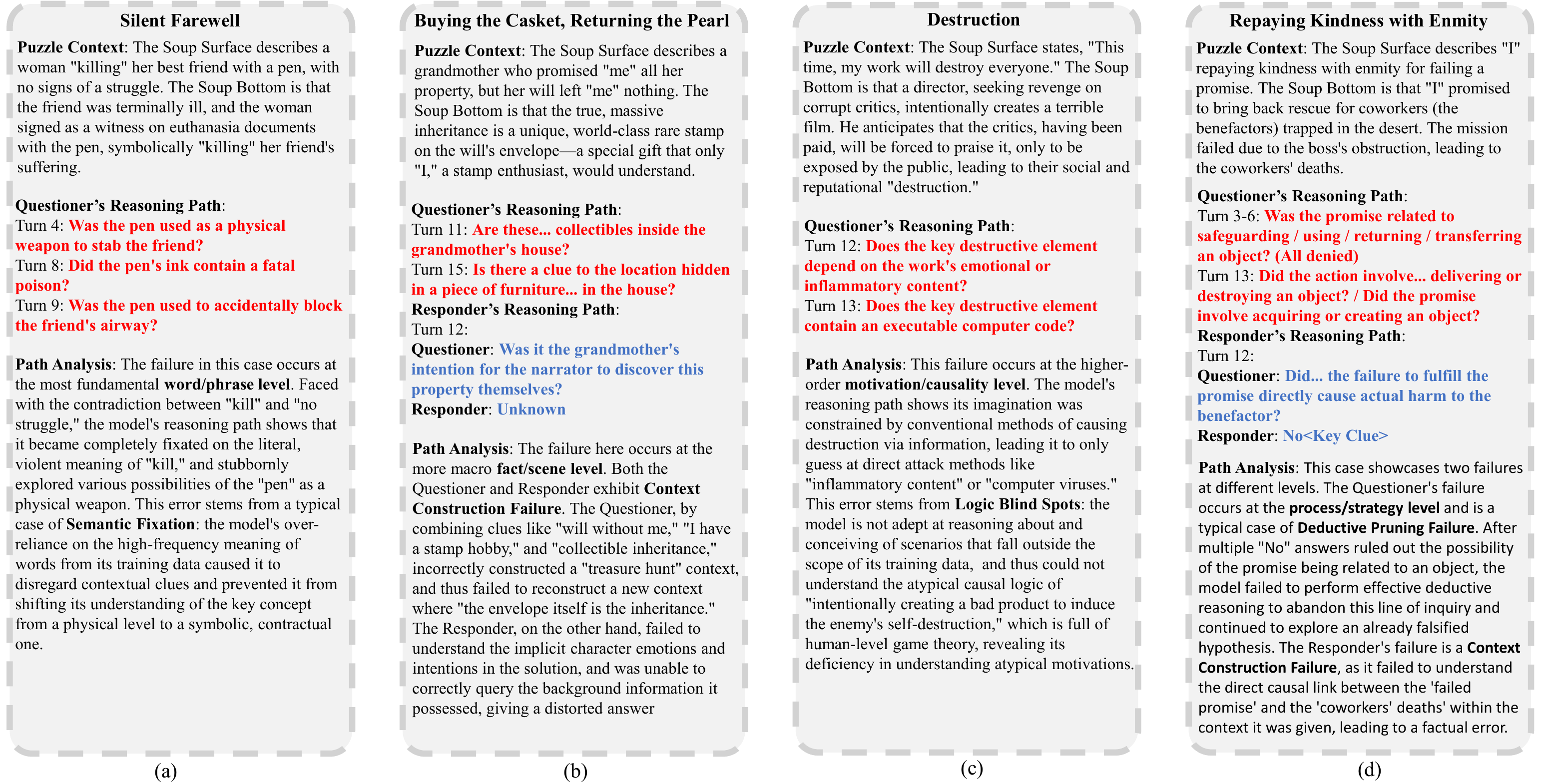}
    
    \caption{Detailed analysis of the four primary failure modes discussed in the main paper: (a) Semantic Fixation, (b) Context Construction Failure, (c) Logic Blind Spots, and (d) Deductive Pruning Failure. Each case provides the puzzle context, the agent's reasoning path, and a path analysis.}
    \label{fig:failure_analysis}
\end{figure*}

This section provides detailed examples for the four typical failure paradigms discussed in the main paper's ``Qualitative and Error Analysis'' section. The complete analysis is presented in Figure \ref{fig:failure_analysis}.

\par\noindent\textbf{Semantic Fixation.}
This is a fundamental failure where the model fixates on the literal, high-frequency meaning of a word, preventing it from understanding symbolic or contextual meanings, even when clues contradict the literal interpretation. For instance, in the ``Silent Farewell'' case, the puzzle presented the seemingly contradictory scenario of a woman ``killing'' her friend with a pen, yet with no signs of a struggle. The model’s reasoning became trapped by the violent connotation of ``kill.'' Its questions demonstrated this fixation, as it explored whether the pen was used for stabbing, contained poison, or blocked an airway. It was unable to shift its understanding to the symbolic, contractual level required to solve the puzzle.

This is a more macro-level error where the agent fails to integrate scattered clues into a coherent and correct model of the situation. This is exemplified in the ``Buying the Casket, Returning the Pearl'' puzzle, where the true inheritance was a rare stamp on a will's envelope. The Questioner agent took individual clues like ``collectible inheritance'' and ``will without me'' and incorrectly constructed a ``treasure hunt'' context, leading it to search for items inside the house. This failure to build the correct context—that the envelope itself was the treasure—derailed its reasoning. The Responder agent also showed this failure by not understanding the grandmother's intent and providing a distorted ``Unknown'' answer.

\par\noindent\textbf{Logic Blind Spots.}
This higher-order failure occurs when the model cannot conceive of atypical motivations or complex, non-linear causality that falls outside of common patterns in its training data. In the ``Destruction'' case, a director seeks to ruin corrupt critics by intentionally making a terrible film they are paid to praise, leading to their reputational ``destruction.'' The model’s imagination was constrained by conventional methods of causing harm via information. Its reasoning path only explored direct attacks like ``inflammatory content'' or ``computer viruses.'' It completely missed the human-level game theory involved, revealing a deficiency in understanding atypical motivations.

\par\noindent\textbf{Deductive Pruning Failure.}
This is a failure in the reasoning process itself, where the agent does not effectively use negative feedback to eliminate incorrect hypotheses and instead continues to explore an already falsified path. The ``Repaying Kindness with Enmity'' case clearly shows this issue. After multiple ``No'' answers from turns 3-6 definitively ruled out that the promise was related to a physical object, the Questioner failed to prune this deductive branch. It continued to pursue this disproven line of inquiry, asking again about an object in turn 13.

\section{Experimental Setup and Evaluation Details}
\label{app:setup}

This section provides a comprehensive overview of the technical details behind our experiments, including software implementation, model configurations, agent hyperparameters, and the specifics of our automated evaluation protocol. All details are reported to ensure full reproducibility of our work.

\subsection{Implementation Details}

\noindent\textbf{Software and Hardware.}
All experiments were conducted in a Python environment, utilizing the asyncio library for concurrent processing of scenarios. LLM interactions were handled via provider-specific Python libraries, primarily through an abstraction layer compatible with the OpenAI API format. No specialized local hardware was required for computation, as all intensive tasks were offloaded to the respective model providers' APIs.

\par\vspace{0.5em}
\noindent\textbf{Model Versions and API Parameters.}
To ensure the stability and reproducibility of our results, all generative calls to the LLM APIs were made with a fixed random seed (seed=42). The temperature is set to 0. Other primary API parameters, such as top\_p, were left at the provider's default values. For tasks requiring structured output (e.g., summary generation, evaluation), the API was instructed to return a JSON object via the response\_format parameter where available. The specific model versions used in our experiments are detailed in Table \ref{tab:model_versions}.

\begin{table}[t]
\centering
\begin{tabularx}{\columnwidth}{lX}
\toprule
\textbf{Model Name} & \textbf{Identifier/Version from Code} \\
\midrule
claude-3.7-sonnet & claude-3-7-sonnet-20250219 \\
gemini-2.5-flash & gemini-2.5-flash-preview-05-20 \\
deepseek-r1 & deepseek-r1 \\
gpt-4o & gpt-4o \\
qwen3-32b & qwen3-32b \\
llama3-8b-instruct & llama3-8b-instruct \\
\bottomrule
\end{tabularx}
\caption{Model Identifiers Used in Experiments.}
\label{tab:model_versions}
\end{table}

\subsection{Mosaic-Agent Hyperparameters}

The behavior of the Mosaic-Agent is governed by several key hyperparameters defined in the execution script. These are listed in Table \ref{tab:agent_hyperparams}. The Meta-cognition module's trigger for re-evaluating the puzzle genre is procedural: it runs if 3 or more new key clues have been discovered since the last check, or if 5 turns have passed without a re-evaluation.

\begin{table}[h]
\centering
\begin{tabularx}{\columnwidth}{X l} 
\toprule
\textbf{Parameter} & \textbf{Value} \\
\midrule
Deliberation Interval ($k$) & 5 turns \\
Max Turns ($N_{\text{max}}$) & 30 \\
EMA Alpha ($\alpha$) & 0.7 \\
Switch Threshold ($\tau_{\text{switch}}$) & 0.1 \\
History Window (for Q-Generation) & 10 turns \\
History Window (for Screening) & 4 turns \\
\bottomrule
\end{tabularx}
\caption{Key Hyperparameters for the Mosaic-Agent.}
\label{tab:agent_hyperparams}
\end{table}

\subsection{Automated Evaluation Protocol Details}
Our automated protocol uses the \texttt{deepseek-r1} model as an LLM-as-a-judge. The process is deterministic, with all evaluation calls also using seed=42.

\par\noindent\textbf{Heuristic Rules for Point Extraction.}
The number of Core Logic Points (N) and Key Details (M) to be extracted from a \textit{soup bottom} is determined adaptively based on its text length and sentence count. For instance, for logic points, a text length under 180 characters yields N=2, while a length over 500 yields N=5. For details, the number M ranges from 3 to 8, adjusted based on both text length and the number of sentences. All values are clipped to a final range of N $\in [2, 5]$ and M $\in [3, 8]$.

\par\noindent\textbf{Semantic Matching and Scoring.}
The model compares a \emph{soup bottom} point against a list of predicted points and return a JSON object containing the best match and a similarity score from 0.0 to 1.0. This score is then calibrated using two thresholds:
\begin{itemize}
    \item \textbf{Validity Threshold:} A match is only considered valid if its similarity score is $\geq 0.5$.
    \item \textbf{High-Confidence Threshold:} If a valid match has a score $\geq 0.8$, its score is normalized to 1.0 to reward strong paraphrases.
\end{itemize}

\par\noindent\textbf{Overall Score Weights:}
The final score is a weighted average of the three metrics:
\begin{itemize}
    \item $w_l$ (Logic): 0.3, $w_d$ (Details): 0.3, $w_c$ (Conclusion): 0.4.
\end{itemize}

\section{Core Prompts for Mosaic-Agent and Evaluation}
\label{app:prompts}

This section contains the core prompts used to guide the behavior of the Mosaic-Agent. Our implementation includes both Chinese and English versions for each prompt; for clarity, only the English versions are presented here. Each prompt is displayed in a separate formatted block.

\subsection{Responder Agent Prompts}

\subsubsection{Answer Generation}
This prompt instructs the Responder to act as the omniscient ``God," comparing the user's question against the \emph{soup bottom} to provide a ``Yes," ``No," or ``Unknown" answer.

\begin{table*}[t!]
\centering 
\fbox{\begin{minipage}{\dimexpr\textwidth-2\fboxsep-2\fboxrule\relax}

\subsection*{Game Background and Role}
You are the judge for a "Situation Puzzle" (or "Lateral Thinking Puzzle," commonly known as "Turtle Soup") game. Your task is to evaluate the \texttt{<Player's Guess>} based on the provided \texttt{<Scenario Description>} (brief story) and \texttt{<Full Explanation>} (complete truth).

\subsection*{Core Judging Principle}
Judge strictly based on the textual content of the \texttt{<Scenario Description>} and \texttt{<Full Explanation>}. Your answer must be one of: "Yes", "No", or "Unknown".

\subsection*{Judging Rules}
\begin{enumerate}[label=\arabic*., itemsep=0.5em, leftmargin=*]
    \item \textbf{Rule as "Yes":}
    \begin{itemize}[nosep, leftmargin=1.5em]
        \item The player's guess aligns with facts, plot points, character actions, or causal relationships explicitly stated in the \texttt{<Scenario Description>} and \texttt{<Full Explanation>}.
        \item The player's guess, although not directly stated, can be derived through direct and necessary logical inference based solely on the information in the \texttt{<Scenario Description>} and \texttt{<Full Explanation>}.
    \end{itemize}

    \item \textbf{Rule as "No":}
    \begin{itemize}[nosep, leftmargin=1.5em]
        \item The player's guess contradicts facts, plot points, character actions, or causal relationships explicitly stated in the \texttt{<Scenario Description>} and \texttt{<Full Explanation>}.
        \item The player's guess can be determined to be false through direct and necessary logical inference based solely on the information.
    \end{itemize}

    \item \textbf{Rule as "Unknown":}
    \begin{itemize}[nosep, leftmargin=1.5em]
        \item The player's guess is not mentioned in either text, and cannot be deduced through any direct, non-speculative logical inference.
        \item Judging the guess would require introducing external common sense or assumptions not present in the text.
    \end{itemize}
\end{enumerate}

\subsection*{Response Requirements}
\begin{itemize}[nosep, leftmargin=*, itemsep=0.3em]
    \item Answer only with "Yes", "No", or "Unknown".
    \item Providing any form of explanation, reasoning, hints, or additional information is prohibited.
\end{itemize}

\subsection*{Special Emphasis}
\begin{itemize}[nosep, leftmargin=*, itemsep=0.3em]
    \item \textbf{Sole Source of Information:} The judgment is based \textbf{only} on the provided texts. Never rely on external knowledge.
    \item \textbf{Limited Inference Allowed:} Only direct, objective logical inference is permissible. Avoid subjective speculation.
\end{itemize}

\subsection*{Puzzle Content}
\par\noindent\texttt{\#\#\# Scenario Description}
\par\noindent\texttt{\{surface\}}
\par\noindent\texttt{\#\#\# Full Explanation}
\par\noindent\texttt{\{bottom\}}
\par\vspace{0.5em}
\par\noindent Now, please judge the following player's guess:

\end{minipage}}
\caption{Prompt for Answer Generation.}
\label{tab:prompt_answer}
\end{table*}

\subsubsection{Key Clue Identification}
This prompt asks the Responder to determine if a given question is semantically related to any of the Key Clue Library for the puzzle.

\begin{table*}[t!]
\centering 
\fbox{\begin{minipage}{\dimexpr\textwidth-2\fboxsep-2\fboxrule\relax}

\subsection*{Task Objective}
Based on the provided scenario description, full explanation, and the key clue library for the current story, please determine if the player's guess pertains to a key clue. Answer directly with Yes/No, without any explanation.

\subsection*{Information}
\par\noindent\texttt{<Scenario Description>}
\par\noindent\texttt{\{surface\}}

\par\vspace{0.5em}
\par\noindent\texttt{<Full Explanation>}
\par\noindent\texttt{\{bottom\}}

\par\vspace{0.5em}
\par\noindent\texttt{<Key Clues>}
\par\noindent\texttt{\{tips\}}

\par\vspace{0.5em}
\par\noindent\texttt{<Player's Question>}
\par\noindent\texttt{\{question\}}

\subsection*{Response Requirements}
\begin{itemize}[nosep, leftmargin=*, itemsep=0.3em]
    \item Analyze whether the player's question touches upon the core causal logic by referencing the \texttt{<Key Clues>} in conjunction with the \texttt{<Full Explanation>}.
    \item Do not answer the player's question itself. Instead, determine if the player's question is a key clue, if it is decisive for deducing the \texttt{<Full Explanation>}, or if it reveals a core event of the scenario or the truth. Does it align with the truth and significantly advance the reasoning process?
    \item Appropriate synonym substitution for the player's question is permissible during analysis.
\end{itemize}

\subsection*{Final Conclusion (Yes/No)}
\textit{Based on the analysis, provide a final, single-word answer: "Yes" or "No".}

\end{minipage}}
\caption{Prompt for Key Clue Identification.}
\label{tab:prompt_is_key}
\end{table*}

\subsection{Questioner Agent Prompts}

\subsubsection{Deliberation Agent}
This set of prompts forms the agent's core reasoning and belief-updating loop.

\par\noindent\textbf{Local Analysis Prompt.} Used after each turn to analyze the immediate implications of the latest question-answer pair.

\begin{table*}[t!]
\centering 
\fbox{\begin{minipage}{\dimexpr\textwidth-2\fboxsep-2\fboxrule\relax}

\subsection*{Task Objective}
Analyze the impact of the new answer on the current understanding of the story. Please perform the analysis based on the following information:

\subsection*{Information Sources}
\par\noindent\texttt{<New Q\&A>}
\par\noindent\texttt{\{answer\}}

\par\vspace{0.5em}
\par\noindent\texttt{<Historical Q\&A>}
\par\noindent\texttt{\{history\_str\}}

\par\vspace{0.5em}
\par\noindent\texttt{<Scenario Description>}
\par\noindent\texttt{\{setup\}}

\subsection*{Analysis Requirements}
Please analyze from the following three aspects:
\begin{itemize}[nosep, leftmargin=*, itemsep=0.3em]
    \item \textbf{New Information Revealed:} Did the answer provide new key information about the scenario description? If so, please explain what the new information is.
    \item \textbf{Knowledge Conflict:} Does the answer conflict with the current understanding or previous information? If so, please specify the point of conflict.
    \item \textbf{Understanding Adjustment:} Based on this answer, how do you suggest adjusting the understanding of the scenario description? Please provide specific adjustment suggestions.
\end{itemize}

\subsection*{Output}
Please list the findings for each aspect separately and organize them to aid subsequent questioning.

\end{minipage}}
\caption{Prompt for Local Analysis.}
\label{tab:prompt_analyze}
\end{table*}

\par\noindent\textbf{Global Deliberation Prompt (Belief State Generation).} Used periodically to synthesize the entire history and all key clues into a structured Belief State ($B_t$).

\begin{table*}[t!]
\centering
\fbox{\begin{minipage}{\dimexpr\textwidth-2\fboxsep-2\fboxrule\relax}

\subsection*{Task Objective}
Based on the provided scenario description, key clue records, historical Q\&A, and the previous understanding, generate a comprehensive and logically rigorous summary aimed at thoroughly explaining all aspects of the entire situation. The summary must include three parts: Detail Summary, Logical Reasoning, and Final Conclusion.

\subsection*{Information Sources}
\par\noindent\texttt{<Scenario Description>}
\par\noindent\texttt{\{surface\}}
\par\vspace{0.5em}
\par\noindent\texttt{<Key Clues>}
\par\noindent\texttt{\{tips\}}
\par\vspace{0.5em}
\par\noindent\texttt{<Historical Q\&A>}
\par\noindent\texttt{\{history\}}
\par\vspace{0.5em}
\par\noindent\texttt{<Previous Understanding>}
\par\noindent\texttt{\{last\_summary\}}

\subsection*{Summary Requirements}
\subsubsection*{1. Detail Summary}
\begin{itemize}[nosep, leftmargin=*, itemsep=0.3em]
    \item \textbf{Goal:} Precisely list all specific points of information crucial for understanding the story's causality, background, or final explanation.
    \item \textbf{Source:} Must comprehensively comb through explicit and implicit information in the \texttt{<Scenario Description>}, \texttt{<Key Clues>}, and \texttt{<Historical Q\&A>}.
    \item \textbf{Requirements:} Each detail should be specific, comprehensive, and objective. Reference \texttt{<Previous Understanding>} to update or supplement details.
\end{itemize}

\subsubsection*{2. Logical Reasoning}
\begin{itemize}[nosep, leftmargin=*, itemsep=0.3em]
    \item \textbf{Goal:} Reveal and describe the core logical chain or underlying pattern that drives the story's development.
    \item \textbf{Focus:} Explain the internal reasons and developmental sequence behind seemingly contradictory or unusual aspects in the \texttt{<Scenario Description>}.
    \item \textbf{Requirements:} The logical chain must be abstract (avoiding specific names/details), coherent, and comprehensive. It should be a general framework applicable to similar stories.
\end{itemize}

\subsubsection*{3. Final Conclusion}
\begin{itemize}[nosep, leftmargin=*, itemsep=0.3em]
    \item \textbf{Goal:} Integrate the Logical Reasoning and Detail Summary to form a fluent, detailed, and fully self-consistent explanation of the story.
    \item \textbf{Construction Method:} Use the framework from Logical Reasoning as a backbone, weaving in specific points from the Detail Summary as evidence.
    \item \textbf{Requirements:} The conclusion must be complete, consistent with all details and logic, fluent, and informative.
\end{itemize}

\subsubsection*{4. Handling Insufficient Information}
\begin{itemize}[nosep, leftmargin=*, itemsep=0.3em]
    \item If information is insufficient, reasonable inferences are permitted to fill in gaps.
    \item Inferred content must be highly relevant and logically self-consistent. Facts derived from inference should be added to the Detail Summary.
\end{itemize}

\subsection*{Output Format}
Strictly output according to the following JSON format. Do not include any JSON comments.
\par\noindent\texttt{\{"details": ["Detail 1", ...], "logic": ["Logic 1", ...], "conclusion": "..."\}}

\end{minipage}}
\caption{Prompt for Global Deliberation / Belief State Generation.}
\label{tab:prompt_summarize}
\end{table*}

\par\noindent\textbf{APS Generation Prompt.} Takes the Belief State as input and instructs the agent to identify doubts and generate the Analysis and Proposal Set.

\begin{table*}[t!]
\centering
\fbox{\begin{minipage}{\dimexpr\textwidth-2\fboxsep-2\fboxrule\relax}

\subsection*{Task Objective}
In the Turtle Soup game, based on the provided \texttt{<Setup>}, \texttt{<Key Clues>}, and \texttt{<Current Summary>}, your task is:
\begin{itemize}[nosep, leftmargin=*, itemsep=0.3em]
    \item \textbf{Identify points of doubt:} Find any vague, contradictory, or insufficiently explained aspects within the \texttt{<Current Summary>}, as well as key elements in the \texttt{<Setup>} that are not yet fully covered. You must quote specific words/phrases to pinpoint these doubts.
    \item \textbf{In-depth analysis and questioning:} For each identified point of doubt:
    \begin{itemize}[nosep, leftmargin=1.5em]
        \item Think deeply about the various different explanations or story developments that might lie behind it.
        \item Based on these thoughts, clearly articulate your analysis.
        \item Propose a specific "yes/no" type question suggestion that effectively explores one of the possibilities from your analysis.
    \end{itemize}
\end{itemize}

\subsection*{Information Sources}
\par\noindent\texttt{<Setup>}
\par\noindent\texttt{\{setup\}}
\par\vspace{0.5em}
\par\noindent\texttt{<Key Clues>}
\par\noindent\texttt{\{clues\}}
\par\vspace{0.5em}
\par\noindent\texttt{<Current Summary>}
\par\noindent\texttt{<Logic>}
\par\noindent\texttt{\{logic\}}
\par\noindent\texttt{<Details>}
\par\noindent\texttt{\{details\}}
\par\noindent\texttt{<Conclusion>}
\par\noindent\texttt{\{conclusion\}}

\subsection*{Output Format}
For each point of doubt, output in the following format:
\begin{itemize}[nosep, leftmargin=*]
    \item "Point of Doubt" + Analysis and Exploration Paths + Question Suggestion
\end{itemize}

\end{minipage}}
\caption{Prompt for Analysis and Proposal Set Generation.}
\label{tab:prompt_advice}
\end{table*}

\subsubsection{Meta-cognition Agent}
This agent uses one of the following six prompts based on the agent's current belief about the puzzle's genre. Each prompt provides a distinct strategic direction for question generation.

\begin{table*}[t!]
\centering
\fbox{\begin{minipage}{\dimexpr\textwidth-2\fboxsep-2\fboxrule\relax}

\subsection*{Task Objective}
You are an experienced Situation Puzzle (Turtle Soup) story classification expert. Your task is to accurately select one type from the predefined list below that best encapsulates the core characteristics of the story, based on the provided scenario description, Q\&A history, and key clues.

\subsection*{Information Sources}
\par\noindent\texttt{<Scenario Description>}
\par\noindent\texttt{\{setup\}}
\par\vspace{0.5em}
\par\noindent\texttt{<Historical Q\&A>}
\par\noindent\texttt{\{history\_str\}}
\par\vspace{0.5em}
\par\noindent\texttt{<Key Clues>}
\par\noindent\texttt{\{clues\_str\}}

\subsection*{Classification Requirements}
Please carefully evaluate all the information above, paying special attention to the following points:
\begin{itemize}[nosep, leftmargin=*, itemsep=0.3em]
    \item \textbf{Scenario Nature:} Does it describe an everyday event, a bizarre phenomenon, a criminal act, a logical puzzle, or something else?
    \item \textbf{Historical Logic:} What direction of reasoning does the Q\&A process reveal? Does it involve specific domain knowledge or unconventional logic?
    \item \textbf{Clue Core:} Which core aspect of the puzzle do the key clues point to? (e.g., motive, physical process, psychological state, information gap, wordplay, etc.)
    \item Based on your comprehensive analysis, please determine which of the following types best fits this story:
    \par \texttt{\{available\_types\}}
\end{itemize}

\subsection*{Output Format}
Your answer must strictly adhere to the format, containing only one type name from the \texttt{<Available Types List>} above. Do not add any explanation or other text. Output the selected type name directly.

\end{minipage}}
\caption{Prompt for Genre Classification.}
\label{tab:prompt_type}
\end{table*}

\begin{table*}[t!]
\centering
\fbox{\begin{minipage}{\dimexpr\textwidth-2\fboxsep-2\fboxrule\relax}

\subsection*{Crime Thriller Type}

\par\noindent\textbf{Category Core:}
This story type focuses on criminal events triggered by human malice (murder, kidnapping, abuse, psychological abnormality, etc.). Supernatural or ghost elements are firmly excluded. The core lies in uncovering the cruel criminal facts hidden behind everyday or bizarre phenomena.

\par\vspace{0.8em}
\noindent\textbf{Common Patterns \& Elements:}
\begin{itemize}[nosep, leftmargin=*, itemsep=0.3em]
    \item \textbf{The "Home" Trap:} Familiar environments (under the bed, next door, items in the house, neighbors) are often the source of danger.
    \item \textbf{Sensory Misdirection:} What is heard (dripping sound $\rightarrow$ blood), seen (blurry figure $\rightarrow$ killer/corpse), or touched (cold $\rightarrow$ corpse/prosthetic) are often disguises or keys pointing to horrific truths.
    \item \textbf{Identity Tricks:} The killer might disguise themselves as a harmless character or exploit identity peculiarities (e.g., a blind person misleading).
    \item \textbf{Weaponization of Everyday Objects:} Ordinary items (food, furniture, dolls, photos) are common carriers of criminal evidence or means of committing crimes.
    \item \textbf{Signature Plot Points:} Often involves gruesome elements like dismemberment, hiding bodies, using/desecrating corpses, imprisonment, etc.
\end{itemize}

\par\vspace{0.8em}
\noindent\textbf{Core Reasoning Directions:}
\begin{itemize}[nosep, leftmargin=*, itemsep=0.3em]
    \item \textbf{Reverse Thinking:} Start from the ending or the core abnormal phenomenon and work backward to deduce the chain of actions.
    \item \textbf{Detail Correlation:} Capture and connect seemingly unrelated details (e.g., "hot weather" + "AC off" + "strange smell" $\rightarrow$ decomposing body).
    \item \textbf{Contradiction Analysis:} Identify illogical or self-contradictory points in the story description (e.g., a "blind person" describing something they "saw").
    \item \textbf{Exclusion \& Focus:} Prioritize ruling out supernatural causes and accidents, focusing thoughts on various possibilities of human crime.
    \item \textbf{Motive Speculation:} Put yourself in the victim's or killer's perspective to consider their behavioral logic and possible criminal motives.
\end{itemize}

\par\vspace{0.8em}
\noindent\textbf{Summary \& Key Tips:}
\par\textit{The key to solving lies in identifying "abnormalities" and "contradictions," piecing together all clue fragments, and making bold but logically grounded guesses. The ultimate goal is to reconstruct a crime story that is logical yet unexpected. Always remind yourself: the answer lies in the dark side of human nature and the harshness of reality.}

\end{minipage}}
\caption{Genre Strategy: Crime Thriller.}
\label{tab:prompt_type_crime}
\end{table*}

\begin{table*}[t!]
\centering
\fbox{\begin{minipage}{\dimexpr\textwidth-2\fboxsep-2\fboxrule\relax}

\subsection*{Mind Maze Type}

\par\noindent\textbf{Category Core:}
This story type focuses on a character's abnormal internal mental world or extreme psychological state. The root of the puzzle stems from the character's own mental illness (hallucinations, delusions, DID), extreme emotions (paranoia, obsession, grief), distorted cognition, or concealed, irrational motives. Supernatural elements are excluded; the emphasis is on exploring the character's inner chaos and suffering. The narrator is often unreliable.

\par\vspace{0.8em}
\noindent\textbf{Common Patterns \& Elements:}
\begin{itemize}[nosep, leftmargin=*, itemsep=0.3em]
    \item \textbf{Inner World Trap:} The source of danger is often the character's own thoughts, memories, or unconscious impulses. The external world may be a trigger or a projection.
    \item \textbf{Perceptual Distortion \& Hallucination:} What the character sees, hears, or feels may not be objective reality but products of their mental state (e.g., seeing non-existent people, hearing command hallucinations).
    \item \textbf{Identity Loss \& Dissociation:} May involve multiple personalities, amnesia, or loss of reality due to trauma. The character might not know who they are.
    \item \textbf{Symbolization of the Mundane:} Ordinary items, places, or dates can become symbolic keys that trigger traumatic memories or compulsive behaviors.
    \item \textbf{Pathological Coping Mechanisms:} Characters might develop extreme ways to cope with trauma (e.g., self-harm, fantasy, turning the deceased into objects).
\end{itemize}

\par\vspace{0.8em}
\noindent\textbf{Core Reasoning Directions:}
\begin{itemize}[nosep, leftmargin=*, itemsep=0.3em]
    \item \textbf{Focus on Internal State:} Prioritize asking about the character's feelings, thoughts, mental condition, and memory reliability. Is there a history of mental illness or trauma?
    \item \textbf{Explore Root Motives:} Dig deep into the psychological driving forces behind actions. Is it fear? Love? Delusion? An uncontrollable impulse?
    \item \textbf{Verify Narrative Reliability:} Always question the narrator's perspective. Could their statements be distorted due to their mental state or memory gaps?
    \item \textbf{Connect Behavior \& State:} View the character's abnormal actions as external manifestations of their internal state. Try to understand "what kind of psychological state would cause someone to do this?"
    \item \textbf{Uncover Symbolic Meaning:} Consider if recurring images or objects hold symbolic meaning, reflecting inner conflicts or trauma.
\end{itemize}

\par\vspace{0.8em}
\noindent\textbf{Summary \& Key Tips:}
\par\textit{The key to solving lies in delving deep into the character's inner world and identifying the "abnormal points" in their perception, cognition, or identity. It requires looking beyond surface behavior to understand the distorted psychological logic behind it. Boldly speculate about the character's mental state, but base inferences on story clues. The answer is often hidden within the maze of the mind, not in external conspiracies.}

\end{minipage}}
\caption{Genre Strategy: Mind Game.}
\label{tab:prompt_type_mind}
\end{table*}

\begin{table*}[t!]
\centering
\fbox{\begin{minipage}{\dimexpr\textwidth-2\fboxsep-2\fboxrule\relax}

\subsection*{Supernatural Fantasy Type}

\par\noindent\textbf{Category Core:}
The solution to this story involves forces that transcend realistic logic. The puzzle's twist often relies on supernatural elements such as ghosts, magic, curses, superpowers (premonition, seeing ghosts), reincarnation, or fantasy/sci-fi settings like time travel or parallel universes. The core lies in revealing unrealistic, mysterious, or bizarre elements and their operating rules.

\par\vspace{0.8em}
\noindent\textbf{Common Patterns \& Elements:}
\begin{itemize}[nosep, leftmargin=*, itemsep=0.3em]
    \item \textbf{Ghostly Manifestations \& Hauntings:} Ghosts or spiritual entities appear due to unfulfilled wishes, revenge, or rituals. They are often associated with specific locations, times, or objects (mirrors, relics).
    \item \textbf{Supernatural Abilities or Curses:} Characters possess or encounter extraordinary abilities (foreseeing death, killing with thoughts) or are influenced by curses, which often bring unexpected consequences.
    \item \textbf{Non-human Entities \& Pacts:} The story involves interactions with demons, spirits, or monsters, usually including deals or pacts that come with ironic twists or huge costs.
    \item \textbf{Spatio-temporal Anomalies \& Cycles:} The story may include time loops, time travel, parallel universes, or characters trapped in cycles of reincarnation or body swapping.
    \item \textbf{Blurred Lines Between Life \& Death:} Characters might continue to be active after death without realizing it, or be able to communicate with the deceased.
    \item \textbf{Folklore \& Taboos:} The story incorporates local legends, folk beliefs, or taboos. Violating taboos is key to triggering abnormal events.
\end{itemize}

\par\vspace{0.8em}
\noindent\textbf{Core Reasoning Directions:}
\begin{itemize}[nosep, leftmargin=*, itemsep=0.3em]
    \item \textbf{Embrace Supernatural Settings:} Abandon purely realistic logic; prioritize considering the possibility of ghosts, magic, and other non-realistic factors.
    \item \textbf{Identify Core Fantasy Elements:} Determine the key supernatural setting. Is it ghosts? Magic? Special abilities?
    \item \textbf{Investigate Operating Rules:} What rules does the supernatural element follow? (e.g., Can ghosts only appear at specific times? What are the side effects of a wish?).
    \item \textbf{Look for Irony \& Twists:} In stories involving wishes or pacts, the ending is often ironic—the wish is granted literally, but the outcome is contrary to intentions.
    \item \textbf{Judge Character Status:} Consider if characters might already be dead, possessed, have unknown superpowers, or be stuck in a time loop.
    \item \textbf{Trace Non-realistic Causality:} When analyzing events, accept that the cause might be supernatural intervention rather than physical or psychological factors.
\end{itemize}

\par\vspace{0.8em}
\noindent\textbf{Summary \& Key Tips:}
\par\textit{The key to solving lies in stepping outside the framework of reality and understanding the specific supernatural setting and its "rules of the game." It requires boldly assuming the existence of fantasy phenomena and making logical deductions within that setting. The goal is to reconstruct a story consistent with its own fantasy logic. Always remind yourself: the answer might lie in the mysterious realms beyond reality.}

\end{minipage}}
\caption{Genre Strategy: Supernatural Fantasy.}
\label{tab:prompt_type_supernatural}
\end{table*}

\begin{table*}[t!]
\centering
\fbox{\begin{minipage}{\dimexpr\textwidth-2\fboxsep-2\fboxrule\relax}

\subsection*{Worldly Vicissitudes Type}

\par\noindent\textbf{Category Core:}
In this story type, the ending is caused by unfortunate accidents, ironic twists of fate, profound misunderstandings, complex emotions (love, hate, guilt, sacrifice), or specific predicaments. The key lies in understanding non-malicious motives, information gaps, or the tricks played by fate. The ending often carries a tragic tone or a sense of absurdity, aiming to evoke reflection on the unpredictability of life.

\par\vspace{0.8em}
\noindent\textbf{Common Patterns \& Elements:}
\begin{itemize}[nosep, leftmargin=*, itemsep=0.3em]
    \item \textbf{Fatal Misunderstandings:} Tragic consequences arising from a child's perspective, lack of information, or sensory impairments.
    \item \textbf{Good Intentions Gone Wrong:} Actions taken out of kindness that lead to disastrous outcomes due to unforeseen factors or coincidences.
    \item \textbf{Ironic Fate \& Cruel Coincidences:} Events unfold ironically, and characters' efforts often backfire.
    \item \textbf{Communication Barriers:} Estrangement, secrets, or information not conveyed in time, leading to tragedy.
    \item \textbf{Environmental \& Occupational Risks:} Dangers inherent in specific environments (hospitals, fire scenes, prisons) or professions (magician, soldier) become the cause of tragedy.
    \item \textbf{Hidden Sacrifices \& Complex Emotions:} The ending involves unknown sacrifices, deep love, or intense guilt that drive seemingly incomprehensible actions.
\end{itemize}

\par\vspace{0.8em}
\noindent\textbf{Core Reasoning Directions:}
\begin{itemize}[nosep, leftmargin=*, itemsep=0.3em]
    \item \textbf{Exclude Simple Attributions:} First, rule out pure malicious murder or supernatural forces; consider accidents or misunderstandings.
    \item \textbf{Investigate Deeper Motives:} Analyze characters' motivations. Consider if the underlying reason could be love, responsibility, fear, or despair, even if actions lead to tragedy.
    \item \textbf{Look for Information Gaps:} Is there information asymmetry between characters? Are judgments based on misinformation or flawed perceptions?
    \item \textbf{Focus on Environment \& Coincidence:} Pay attention to the story's setting, time, weather, and any seemingly unrelated coincidences.
    \item \textbf{Understand Irony \& Helplessness:} Try to interpret the story from the perspective of fate's cruelty or things going contrary to intentions.
    \item \textbf{Consider Hidden Background:} Is there unmentioned background information (like past experiences, physical condition, social environment) that could explain the events?
\end{itemize}

\par\vspace{0.8em}
\noindent\textbf{Summary \& Key Tips:}
\par\textit{The key to solving lies in recognizing that the sources of tragedy are often interwoven with accidents, misunderstandings, complex emotions, and environmental factors. It requires empathy to understand the potential non-malicious motives behind characters' actions. The goal is to reconstruct a tragic story that is logically sound and emotionally thought-provoking. Always remind yourself: the answer might be hidden in fate's jokes and human fragility.}

\end{minipage}}
\caption{Genre Strategy: Constant Change.}
\label{tab:prompt_type_change}
\end{table*}

\begin{table*}[t!]
\centering
\fbox{\begin{minipage}{\dimexpr\textwidth-2\fboxsep-2\fboxrule\relax}

\subsection*{Logic and Cleverness Type}

\par\noindent\textbf{Category Core:}
The focus of this story is on the puzzle-solving process and the cleverness of the ending. The twist often relies on breaking conventional logic, shifting perspectives (like the narrator's identity), or wordplay. The ending typically reveals a rational, sometimes ordinary or touching truth behind a seemingly bizarre phenomenon, aiming for an "aha!" moment.

\par\vspace{0.8em}
\noindent\textbf{Common Patterns \& Elements:}
\begin{itemize}[nosep, leftmargin=*, itemsep=0.3em]
    \item \textbf{Perspective Shift \& Identity Misdirection:} The narrator is not the expected person or creature (e.g., an animal, an object, a game character), or an observer misjudges due to a limited viewpoint.
    \item \textbf{Contextual Misdirection:} Events that seem thrilling or criminal are actually ordinary situations from daily life, games, or misunderstandings.
    \item \textbf{Wordplay \& Literal Interpretation:} Using puns, ambiguities, or the literal meaning of words to create confusion.
    \item \textbf{Breaking Conventional Thinking:} The solution depends on breaking ingrained stereotypes about specific behaviors or cause-and-effect relationships.
    \item \textbf{Hidden Information \& Key Details:} A seemingly insignificant detail (a physical characteristic, an object, a rule) is the key to solving the puzzle.
    \item \textbf{Warm Twist \& Unexpected Emotion:} Revealing hidden goodwill, love, or a touching truth behind seemingly negative circumstances.
\end{itemize}

\par\vspace{0.8em}
\noindent\textbf{Core Reasoning Directions:}
\begin{itemize}[nosep, leftmargin=*, itemsep=0.3em]
    \item \textbf{Question Narrator/Perspective Identity:} Who is "I"? Are they human? An animal? An object? Does the observer have all the information?
    \item \textbf{Seek Reasonable Explanation, Not Malice:} Prioritize ruling out crime and the supernatural. What is the simplest explanation that aligns with everyday logic?
    \item \textbf{Pay Attention to Puns \& Ambiguity:} Do keywords in the story have multiple meanings? Are there homophones or wordplay involved?
    \item \textbf{Challenge Fixed Assumptions:} Must character behavior always conform to common sense? Is there a counter-intuitive yet logical explanation?
    \item \textbf{Dig Deep into Abnormal Details:} Why was this specific color/number/object mentioned? What hidden information does it imply?
    \item \textbf{Consider Positive Motives:} Beneath seemingly negative descriptions, could there be positive, touching, or well-intentioned possibilities?
\end{itemize}

\par\vspace{0.8em}
\noindent\textbf{Summary \& Key Tips:}
\par\textit{The key to solving lies in breaking mental fixedness and engaging in multi-angled, unconventional thinking. It requires sensitivity to language, skill in capturing details, and the courage to challenge first impressions. The goal is to find that clever, logically self-consistent, often smile-inducing "brain teaser" answer. Always remind yourself: the truth might be hidden in the most unexpectedly simple logic or perspective shift.}

\end{minipage}}
\caption{Genre Strategy: Clever Logic.}
\label{tab:prompt_type_logic}
\end{table*}

\begin{table*}[t!]
\centering
\fbox{\begin{minipage}{\dimexpr\textwidth-2\fboxsep-2\fboxrule\relax}

\par\noindent\textbf{Goal:} The type of this story is currently unknown. The primary task is preliminary information gathering to quickly grasp the core elements, laying the foundation for subsequent reasoning and accurate type determination. Please prioritize using the following Situation Puzzle questioning strategies:

\begin{enumerate}[label=\textbf{\arabic*.}, wide, labelwidth=!, labelindent=0pt, itemsep=0.5em]
    \item \textbf{Basic Qualification \& Exclusion (Primary Probe Points):}
    \begin{itemize}[nosep, leftmargin=1.5em]
        \item \textit{Authentic/Variant Judgment:} Does the story strictly adhere to real-world logic, or does it contain non-realistic elements such as supernatural, sci-fi, or fantasy?
        \item \textit{Character/Object Nature:} Are all entities mentioned "human"? Do animals, objects, ghosts, or other non-human entities play key roles?
    \end{itemize}

    \item \textbf{Core Information Dimension Probing:}
    \par Based on the available information, proactively probe the following key dimensions. Choose the direction most likely to reveal the truth for questioning:
    \begin{itemize}[nosep, leftmargin=1.5em]
        \item \textit{Character Elements:} Number, gender, occupation, and relationships of key characters?
        \item \textit{Event Elements:} Is it clear someone has died? What was the manner of death? Does the story involve multiple time periods?
        \item \textit{Item/Environment Elements:} Do mentioned items have special significance or quantity?
    \end{itemize}

    \item \textbf{Combined Basic Analysis:}
    \par While gathering information, conduct deeper analysis based on known clues:
    \begin{itemize}[nosep, leftmargin=1.5em]
        \item \textit{Background \& Motive:} Analyze the story's environment and time period, and infer motives behind characters' actions.
        \item \textit{Key Events \& Details:} Pay attention to anomalies, contradictions, or deliberately emphasized details.
        \item \textit{Potential Conflicts \& Perspectives:} Explore unspoken conflicts, information gaps, or different perspectives among characters.
    \end{itemize}
\end{enumerate}

\par\vspace{0.8em}
\par\noindent\textbf{Instruction:} Based on the strategies above, analyze the current scenario description and known information. Select the direction that most urgently needs confirmation to efficiently obtain key information, verify core hypotheses, or eliminate incorrect paths.

\end{minipage}}
\caption{Genre Strategy: Default.}
\label{tab:prompt_type_default}
\end{table*}

\subsubsection{Action Formulation Agent}
These prompts are responsible for generating and selecting the agent's next question.

\par\noindent\textbf{Candidate Question Generation Prompt.} Synthesizes all available analyses and strategic guidance to generate three diverse candidate questions.

\begin{table*}[t!]
\centering
\fbox{\begin{minipage}{\dimexpr\textwidth-2\fboxsep-2\fboxrule\relax}

\subsection*{Task Objective}
You are a question-guiding expert for a "Situation Puzzle" (Turtle Soup) game. Your responsibility is to generate optimal questioning strategies based on the game's progress, using "Yes/No" questions to gradually unveil the puzzle's truth.

\subsection*{Question Rules}
\begin{itemize}[nosep, leftmargin=*, itemsep=0.3em]
    \item Must use "Yes/No" interrogative sentences.
    \item Compound questions (e.g., "Is it A and B?") are prohibited.
\end{itemize}

\subsection*{Task Requirements}
Please generate 3 different "Yes/No" questions. Each question must comply with the rules and strategies above, offering diverse exploration directions. Ensure that:
\begin{itemize}[nosep, leftmargin=*, itemsep=0.3em]
    \item At least one question is based on new information from the \texttt{<Question Suggestions>} or \texttt{<Previous Q\&A Analysis>}.
    \item At least one question focuses on specific events or details (e.g., location, action, object).
    \item Questions explore new dimensions (e.g., time, emotion) to ensure diversity.
\end{itemize}

\subsection*{Question Strategy}
\begin{enumerate}[label=\arabic*., itemsep=0.5em, leftmargin=*]
    \item \textbf{Prioritization:}
    \begin{itemize}[nosep, leftmargin=1.5em]
        \item \textit{Ambiguity First:} Address vague details and logical gaps.
        \item \textit{Utilize New Knowledge:} Adjust direction based on new information from the \texttt{<Previous Q\&A Analysis>}.
        \item \textit{Layered Questioning:} After identifying a broad key point, shift to asking about specific details.
    \end{itemize}
    \item \textbf{Exploration Directions:}
    \begin{itemize}[nosep, leftmargin=1.5em]
        \item Explore story framework elements (time, place, characters, cause, etc.).
        \item If "Unknown" is answered for 2 consecutive rounds, immediately switch direction.
        \item Refer to template guidance from the \texttt{<Template Guidance>}.
    \end{itemize}
    \item \textbf{Inference and Empathy:}
    \begin{itemize}[nosep, leftmargin=1.5em]
        \item Infer story development based on existing clues, paying attention to how character emotions drive events.
        \item Prioritize exploring potential emotional conflicts or motivations.
    \end{itemize}
\end{enumerate}

\subsection*{Current Game State}
\par\noindent\texttt{<Scenario Description>}
\par\noindent\texttt{\{surface\}}
\par\vspace{0.5em}
\par\noindent\texttt{<Question Suggestions>}
\par\noindent\texttt{\{advice\}}
\par\vspace{0.5em}
\par\noindent\texttt{<Previous Q\&A Analysis>}
\par\noindent\texttt{\{answer\_analysis\}}
\par\vspace{0.5em}
\par\noindent\texttt{<Template Guidance>}
\par\noindent\texttt{\{type\_instruction\}}
\par\vspace{0.5em}
\par\noindent\texttt{<Recent History Record>}
\par\noindent\texttt{\{history\}}

\subsection*{Output Format}
Please generate the questions directly without any explanation, returning them in the following format:
\par\vspace{0.5em}
\par\noindent\texttt{1. [Question 1]?}
\par\noindent\texttt{2. [Question 2]?}
\par\noindent\texttt{3. [Question 3]?}

\end{minipage}}
\caption{Prompt for Candidate Question Generation.}
\label{tab:prompt_question}
\end{table*}

\par\noindent\textbf{Optimal Question Screening Prompt.} Instructs the agent to act as a critic, evaluating the three candidates against the full history and a blacklist to select the single best question to ask next.

\begin{table*}[t!]
\centering
\fbox{\begin{minipage}{\dimexpr\textwidth-2\fboxsep-2\fboxrule\relax}

\subsection*{Task Objective}
You are a Question Selection Expert for a "Situation Puzzle" (Turtle Soup) game. Your responsibility is to choose the optimal question from multiple candidates to advance the game process and reveal the puzzle's truth.

\subsection*{Current Game State}
\par\noindent\texttt{<Scenario Description>}
\par\noindent\texttt{\{surface\}}
\par\vspace{0.5em}
\par\noindent\texttt{<Question Suggestions>}
\par\noindent\texttt{\{advice\}}
\par\vspace{0.5em}
\par\noindent\texttt{<Previous Q\&A Analysis>}
\par\noindent\texttt{\{answer\_analysis\}}
\par\vspace{0.5em}
\par\noindent\texttt{<Recent History Record>}
\par\noindent\texttt{\{history\}}
\par\vspace{0.5em}
\par\noindent\texttt{<Asked Questions>}
\par\noindent\texttt{\{history\_questions\}}

\subsection*{Candidate Questions}
\begin{enumerate}[nosep, leftmargin=*]
    \item \texttt{\{question1\}}
    \item \texttt{\{question2\}}
    \item \texttt{\{question3\}}
\end{enumerate}

\subsection*{Error Prevention and Selection Criteria}
\begin{itemize}[nosep, leftmargin=*, itemsep=0.3em]
    \item \textbf{Avoid Redundancy:} The question must not be repetitive or highly similar to existing questions in \texttt{<Asked Questions>}.
    \item \textbf{Focus on Key Points:} Prioritize questions targeting suggestions in \texttt{<Question Suggestions>} or new information/conflicts identified in \texttt{<Previous Q\&A Analysis>}.
    \item \textbf{Dynamic Adjustment:} If the last two rounds in \texttt{<Recent History Record>} were both answered "Unknown", prioritize questions exploring entirely new directions.
    \item \textbf{Blacklist:} Avoid directions that have already been ruled out. Avoid asking questions identical or similar to those on the blacklist below:
    \par \texttt{\{blacklist\}}
    \item \textbf{Advance Reasoning:} Select the question most likely to reveal new clues or resolve logical flaws.
\end{itemize}

\subsection*{Output Requirements}
\begin{itemize}[nosep, leftmargin=*, itemsep=0.3em]
    \item Return the full text of the selected question.
    \item Ensure the output question is identical to one of the candidate questions, with no changes in wording.
\end{itemize}

\subsection*{Note}
If none of the questions meet the criteria, select the first question by default.
\par\vspace{1em}
\par\noindent Please return the best question for this round directly:

\end{minipage}}
\caption{Prompt for Optimal Question Screening.}
\label{tab:prompt_select_question}
\end{table*}

\subsection{Evaluation Prompts}
These prompts are used in our automated evaluation protocol to score the agent's performance.

\subsubsection{\emph{soup bottom} Point Extraction}
This prompt instructs the LLM to read the full solution of a puzzle and decompose it into a structured set of Core Logic Points and Key Details.

\begin{table*}[t!]
\centering
\fbox{\begin{minipage}{\dimexpr\textwidth-2\fboxsep-2\fboxrule\relax}

\subsection*{Task Objective}
Please extract the key logical relationships and detailed information from the following story's true solution \texttt{<Full Explanation>}.

\subsection*{Requirements}

\subsubsection*{[Logical Relationships]}
\begin{itemize}[nosep, leftmargin=*, itemsep=0.3em]
    \item \textbf{Objective:} To reveal and describe the core logical chain or underlying pattern driving the story's development.
    \item \textbf{Quantity:} Please extract around \texttt{\{N\_LOGIC\_POINTS\}} key logical steps.
    \item \textbf{Extraction Requirements:}
    \begin{itemize}[nosep, leftmargin=1.5em]
        \item \textit{Abstraction:} Extract the story's general causal chain. Focus on "how" and "why" events occur. \textbf{Absolutely avoid} using any specific character names, places, times, or other details.
        \item \textit{Coherence:} Construct a complete, unbroken logical chain from cause to effect.
        \item \textit{Comprehensiveness:} The logical chain must explain all major plot points and turning factors.
    \end{itemize}
\end{itemize}

\subsubsection*{[Detailed Information]}
\begin{itemize}[nosep, leftmargin=*, itemsep=0.3em]
    \item \textbf{Objective:} To accurately list all specific pieces of information crucial for understanding the story.
    \item \textbf{Quantity:} Please extract around \texttt{\{M\_DETAIL\_POINTS\}} key pieces of detailed information.
    \item \textbf{Extraction Requirements:}
    \begin{itemize}[nosep, leftmargin=1.5em]
        \item \textit{Specificity:} Extract concrete scenes, characters, objects, times, or facts. Avoid vague descriptions.
        \item \textit{Independence:} List each key piece of information separately.
        \item \textit{Comprehensiveness:} Ensure no crucial details are omitted.
    \end{itemize}
\end{itemize}

\par\vspace{0.5em}
\par\noindent\texttt{<Full Explanation>}
\par\noindent\texttt{\{bottom\}}

\subsection*{Output Format}
Please output directly in the following format, without explanations:
\par\vspace{0.5em}
\par\noindent\texttt{[Logical Relationships]}
\par\noindent\texttt{- Logic 1:}
\par\noindent\texttt{- Logic 2:}
\par\noindent\texttt{- ...}
\par\vspace{0.5em}
\par\noindent\texttt{[Detailed Information]}
\par\noindent\texttt{- Detail 1:}
\par\noindent\texttt{- Detail 2:}
\par\noindent\texttt{- ...}

\end{minipage}}
\caption{Prompt for \emph{soup bottom} Point Extraction.}
\label{tab:prompt_extract}
\end{table*}

\subsubsection{Semantic Similarity Scoring}
This prompt asks the LLM to compare a single \emph{soup bottom} point against a list of predicted points and return a JSON object with the best match and a similarity score.

\begin{table*}[t!]
\centering
\fbox{\begin{minipage}{\dimexpr\textwidth-2\fboxsep-2\fboxrule\relax}
\small 

\subsection*{Task Objective}
\begin{itemize}[nosep, leftmargin=*, itemsep=0.3em]
    \item \textbf{Task Type:} One-to-Many Scoring \& Matching
    \item \textbf{Input:} 1. \texttt{<Base Statement>}; 2. \texttt{<List of Predicted Statements>} (numbered)
    \item \textbf{Core Instructions:}
    \begin{itemize}[nosep, leftmargin=1.5em]
        \item For each item in the \texttt{<List of Predicted Statements>}, independently calculate a semantic similarity score from 0.0 to 1.0 against the \texttt{<Base Statement>}.
        \item From all calculated scores, identify the single highest score and its corresponding "best match".
    \end{itemize}
    \item \textbf{Decision Threshold:} 0.5. Used to determine if the best match constitutes a valid match.
\end{itemize}

\subsection*{Core Semantic Judgment Criteria}
When generating scores, strictly adhere to this principle: Similarity focuses on the potential meaning, which is the core event or outcome. It does not require identical phrasing; you should ignore differences in expression, background information, or minor details. A score of 1.0 represents identical core meaning; 0.0 represents completely different.

\subsection*{Statements}
\par\noindent\texttt{<Base Statement>}
\par\noindent\texttt{\{ground\_truth\}}
\par\vspace{0.5em}
\par\noindent\texttt{<List of Predicted Statements>}
\par\noindent\texttt{\{predicted\_list\}}

\subsection*{Output Requirements}
The response body must and can only be a JSON object, containing no other text. The structure of this JSON object must strictly follow one of the two preset schemas below.

\par\vspace{0.5em}
\par\noindent\textbf{Schema A: Match Successful} (when the best match's score is $\geq$ 0.5)
\par\noindent\texttt{\{}
\par\noindent\texttt{\ \ "best\_match\_index": <Integer index of the best matching statement>,}
\par\noindent\texttt{\ \ "best\_match\_score": <Precise float score of that statement>}
\par\noindent\texttt{\}}

\par\vspace{0.8em}
\par\noindent\textbf{Schema B: Match Failed} (when the best match's score is $<$ 0.5)
\par\noindent\texttt{\{}
\par\noindent\texttt{\ \ "best\_match\_index": null,}
\par\noindent\texttt{\ \ "best\_match\_score": <The actual highest score found>}
\par\noindent\texttt{\}}

\end{minipage}}
\caption{Prompt for Semantic Similarity Scoring.}
\label{tab:prompt_similarity}
\end{table*}

\end{document}